\documentclass{article} %
\usepackage{arxiv_conference,times}

\usepackage{amsmath,amsfonts,bm}

\newcommand{\ba}[1]{\begin{align}#1\end{align}}

\newcommand{\given}{\,|\,}

\def\Figref#1{Figure~\ref{#1}}

\def\eqref#1{equation~\ref{#1}}
\def\Eqref#1{Equation~\ref{#1}}

\def\1{\bm{1}}

\def\rvepsilon{{\mathbf{\epsilon}}}
\def\rvtheta{{\mathbf{\theta}}}

\def\rvm{{\mathbf{m}}}

\def\rvx{{\mathbf{x}}}

\def\rvz{{\mathbf{z}}}

\DeclareMathAlphabet{\mathsfit}{\encodingdefault}{\sfdefault}{m}{sl}
\SetMathAlphabet{\mathsfit}{bold}{\encodingdefault}{\sfdefault}{bx}{n}

\newcommand{\E}{\mathbb{E}}

\DeclareMathOperator*{\argmin}{arg\,min}

\usepackage[utf8]{inputenc} %
\usepackage[T1]{fontenc}    %
\usepackage{hyperref}       %
\usepackage{url}            %
\usepackage{booktabs}       %
\usepackage{amsfonts}       %
\usepackage{nicefrac}       %
\usepackage{microtype}      %
\usepackage{xcolor}         %
\usepackage{bbold}
\usepackage{caption}
\usepackage{subcaption}
\usepackage[ruled,longend, algo2e]{algorithm2e}
\usepackage{algorithm}
\usepackage{algorithmic}
\usepackage{hhline}
\usepackage{multirow}
\usepackage{stmaryrd}
\usepackage{enumitem,pifont}
\usepackage{etoolbox}
\usepackage{wrapfig}
\usepackage{xcolor}  
\usepackage{graphicx}
\usepackage{multirow}
\usepackage{array}
\usepackage{tabularx}
\usepackage{makecell}
\usepackage{colortbl}
\usepackage{wrapfig}
\usepackage{pifont}
\newcommand{\cmark}{\ding{51}}%
\newcommand{\xmark}{\ding{55}}%

\definecolor{codegreen}{rgb}{0,0.3,0.6}
\definecolor{codegray}{rgb}{0.5,0.5,0.5}
\definecolor{codepurple}{rgb}{0.58,0,0.82}
\definecolor{backcolour}{rgb}{0.95,0.95,0.92}
\definecolor{orange}{rgb}{1,0.5,0}
\definecolor{mydarkblue}{rgb}{0,0.08,0.45}

\newcommand{\RN}[1]{%
	\textup{\lowercase\expandafter{\it \romannumeral#1}}%
}

\usepackage{listings}
\lstdefinestyle{mystyle}{
    basicstyle=\tiny,
    commentstyle=\color{codegreen},
    keywordstyle=\color{magenta},
    numberstyle=\tiny\color{codegray},
    stringstyle=\color{codepurple},
    basicstyle=\fontsize{8.5}{9}\selectfont\ttfamily,
    breakatwhitespace=false,         
    breaklines=true,                 
    captionpos=b,                    
    keepspaces=true,                 
    numbers=none,
    citecolor=mydarkblue,
    numbersep=5pt,                  
    showspaces=false,                
    showstringspaces=false,
}

\newcommand{\rev}[1]{\textcolor{black}{#1}} 

\title{Learning Stackable and Skippable LEGO Bricks for Efficient, Reconfigurable, and Variable-Resolution
Diffusion Modeling}

\author{Huangjie Zheng$^{1,2}$,~~Zhendong Wang$^{1}$,~~Jianbo Yuan$^{2}$,~~Guanghan Ning$^{2}$\\
\textbf{Pengcheng He$^{3}$,~~
Quanzeng You$^{2}$,~~ Hongxia Yang$^{2}$, ~~ Mingyuan Zhou$^{1}$}\\
$^{1}$The University of Texas at Austin, $^{2}$ByteDance Inc., $^{3}$Microsoft Azure AI
\\
\texttt{\{huangjie.zheng,zhendong.wang\}@utexas.edu}
\\
\texttt{\{jianbo.yuan,guanghan.ning,quanzeng.you,hx.yang\}@bytedance.com}
\\
\texttt{herbert.he@gmail.com, mingyuan.zhou@mccombs.utexas.edu}
}

\iclrfinalcopy %
\begin{document}

\maketitle

\begin{abstract}

Diffusion models excel at generating photo-realistic images but come with significant computational costs in both training and sampling. While various techniques address these computational challenges, a less-explored issue is designing an efficient and adaptable network backbone for iterative refinement. Current options like U-Net and Vision Transformer often rely on resource-intensive deep networks and lack the flexibility needed for generating images at variable resolutions or with a smaller network than used in training.
This study introduces LEGO bricks, which seamlessly integrate Local-feature Enrichment and Global-content Orchestration. These bricks can be stacked to create a test-time reconfigurable diffusion backbone, allowing selective skipping of bricks to reduce sampling costs and generate higher-resolution images than the training data. LEGO bricks enrich local regions with an MLP and transform them using a Transformer block while maintaining a consistent full-resolution image across all bricks. Experimental results demonstrate that LEGO bricks enhance training efficiency, expedite convergence, and facilitate variable-resolution image generation while maintaining strong generative performance. Moreover, LEGO significantly reduces sampling time compared to other methods, establishing it as a valuable enhancement for diffusion models. Our code and project page are available at \url{https://jegzheng.github.io/LEGODiffusion}.

\end{abstract}

\section{Introduction}\label{sec:intro}

Diffusion models, also known as score-based generative models, have gained significant traction in various domains thanks to their proven effectiveness 
in generating high-dimensional data
and 
offering simple implementation
\citep{sohl2015deep,song2019generative,song2020improved,ho2020denoising,song2021scorebased}. Decomposing data generation into simple denoising tasks across evolving timesteps with varying noise levels,  they have played a pivotal role in driving progress in a wide range of fields, with a particular emphasis on their contributions to image generation \citep{dhariwal2021diffusion,nichol2021glide,ramesh2022hierarchical,saharia2022photorealistic,rombach2022high}.

\begin{figure}[t]
    \centering
    \includegraphics[width=\textwidth]{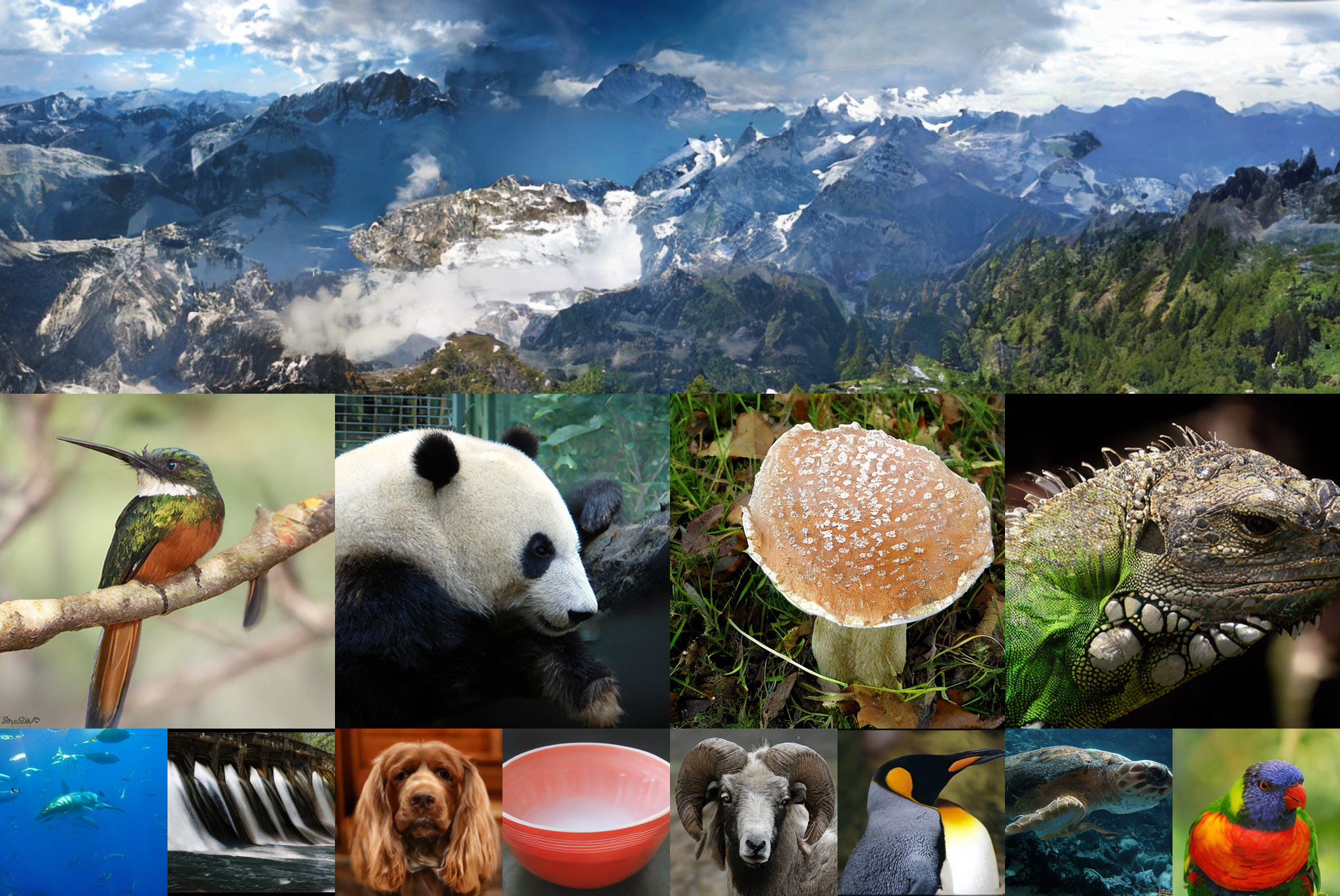}\vspace{-8pt}
    \caption{\small Visualization of LEGO-generated images. Top Row: $2048 \times 600$ panorama image sample, from the model trained on ImageNet $256\times256$. Middle Row: $512\times512$ image samples, trained on ImageNet $512\times512$. Bottom Row: $256\times256$ image samples, trained on ImageNet $256\times256$.}\vspace{-9pt}
    \label{fig:imagenet-vis}\vspace{-6mm}
\end{figure}

Despite these advancements, diffusion models continue to face a significant challenge: their substantial computational requirements, not only during training but also during sampling. In sampling, diffusion models typically demand a large number of functional evaluations (NFE) to simulate the reverse diffusion process.
To tackle this challenge, various strategies have been proposed, including methods aimed at expediting sampling through approximated reverse steps \citep{song2020denoising,zhang2022fast,lu2022dpm,karras2022edm}, as well as approaches that integrate diffusion with other generative models \citep{zheng2022truncated,pandey2022diffusevae}. 

Another clear limitation is that the network used in sampling must generally be the same as that used in training, unless progressive distillation techniques are applied \citep{salimans2022progressive,song2023consistency}. With or without distillation, the network used for sampling typically lacks reconfigurability and would require retraining if a smaller memory or computational footprint is desired.

Beside the computing-intensive and rigid sampling process, training diffusion models also entails a considerable number of iterations to attain convergence to an acceptable checkpoint. This requirement arises from the model's need to learn how to predict the mean of clean images conditioned on noisy inputs, which exhibit varying levels of noise. This concept can be concluded from either the Tweedie's formula \citep{robbins1992empirical,efron2011tweedie}, as discussed in \citet{luo2022understanding} and \citet{chung2022improving},  or the Bregman divergence \citep{banerjee2005clustering}, as illustrated in \citet{zhou2023beta}. This mechanism imposes a significant computational burden, particularly when dealing with larger and higher dimensional datasets.
To mitigate the training cost, commonly-used approaches include training the diffusion model in a lower-dimensional space and then recovering the high-dimensional data using techniques such as pre-trained auto-encoders \citep{rombach2022high,li2022diffusion} or cascaded super-resolution models \citep{ho2022cascaded,saharia2022photorealistic}. An alternative strategy involves training diffusion models at the patch-level input \citep{luhman2022improving}, requiring additional methods to ensure generated samples maintain a coherent global structure. These methods may include incorporating full-size images periodically \citep{wang2023patch}, using encoded or downsampled semantic features \citep{ding2023patched,arakawa2023memory}, or employing masked transformer prediction \citep{zheng2023fast,gao2023masked}.

Despite numerous efforts to expedite diffusion models, a significant but relatively less-explored challenge remains in designing an efficient and flexible network backbone for iterative refinement during both training and sampling.
Historically, the dominant choice for a backbone has been based on U-Net \citep{ronneberger2015unet}, but recently, an alternative backbone based on the Vision Transformer (ViT) \citep{dosovitskiy2021an} has emerged as a compelling option.
However, it's important to note that both U-Net and ViT models still require the inclusion of deep networks with a substantial number of computation-intensive convolutional layers or Transformer blocks to achieve satisfactory performance. 
Moreover, it's important to highlight that neither of these options readily allows for network reconfiguration during sampling, and generating images at higher resolutions than the training images frequently poses a significant challenge.

Our primary objective is to introduce the ``LEGO brick,'' a fundamental network unit that seamlessly integrates two essential components: \textbf{L}ocal-feature \textbf{E}nrichment and \textbf{G}lobal-content \textbf{O}rchestration. 
These bricks can be vertically stacked to form the reconfigurable backbone of a diffusion model that introduces spatial refinement within each timestep. This versatile backbone not only enables selective skipping of LEGO bricks for reduced sampling costs but also capably generates images at resolutions significantly higher than the training set.

A LEGO diffusion model is assembled by stacking a series of LEGO bricks, each with different input sizes. Each brick processes local patches whose dimensions are determined by its individual input size, all aimed at refining the spatial information of the full-resolution image.
In the process of ``Local-feature Enrichment,'' a specific LEGO brick takes a patch that matches its input size ($e.g.$, a $16\times 16$ noisy patch) along with a prediction from the preceding brick. This patch-level input is divided into non-overlapping local receptive fields ($e.g.$, four $8\times 8$ patches), and these are represented using local feature vectors projected with a ``token'' embedding layer.
Following that, ``Global-content Orchestration'' involves using Transformer blocks~\citep{vaswani2017attention,dosovitskiy2021an}, comprising multi-head attention and MLP layers, to re-aggregate these local token-embedding feature vectors into a spatially refined output that matches the input size. This approach results in an efficient network unit achieved through MLP mixing that emphasizes local regions while maintaining a short attention span ($e.g.$, a sequence of four token-embedding vectors). Importantly, each LEGO brick is trained using sampled input patches rather than entire images, significantly reducing the computational cost associated with the model's forward pass.
This approach not only enables the flexible utilization of LEGO bricks of varying sizes during training but also facilitates a unique reconfigurable architecture during generation. Furthermore, it empowers the model with the capability to generate images at significantly higher resolutions than those present in the training dataset, as illustrated in \Figref{fig:imagenet-vis}.

\begin{figure}
  \centering
  \vspace{-15pt} 
  \includegraphics[width=\textwidth]{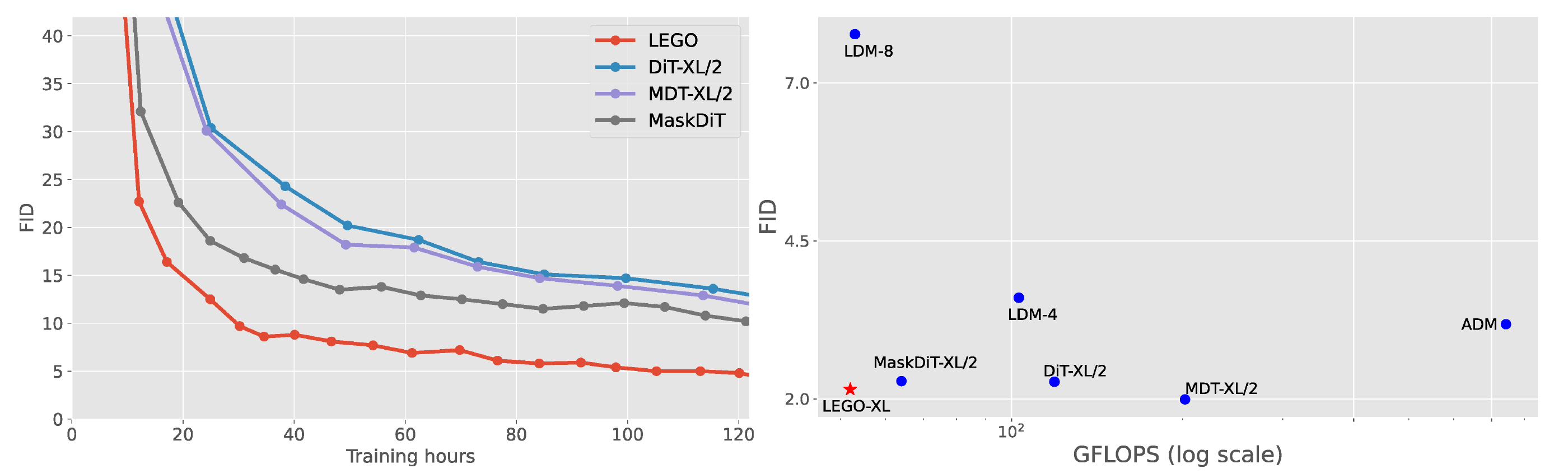}\vspace{-4mm}
  \caption{\small With classifier-free guidance, we present the following: Left Panel: A comparison of convergence, measured with FID  versus training time. \rev{Right Panel: A comparison of the computation cost measured with FID versus training FLOPs.} These experiments were conducted using eight NVIDIA A100 GPUs on ImageNet ($256 \times 256$), with a batch size of 256.}\label{fig:conv_flops}
  \vspace{-6mm}
\end{figure}

In summary, the stackable and skippable LEGO bricks possess several noteworthy characteristics in managing computational costs in both training and sampling, offering a flexible and test-time reconfigurable backbone, and facilitating variable-resolution generation.
Our experimental evaluations, as shown in Figures~\ref{fig:imagenet-vis} and \ref{fig:conv_flops}, clearly demonstrate that LEGO strikes a good balance between efficiency and performance across challenging image benchmarks.
LEGO significantly enhances training efficiency, as evidenced by reduced FLOPs, faster convergence, and shorter training times, all while maintaining a robust generation performance. These advantages extend seamlessly into the sampling phase, where LEGO achieves a noteworthy 60\% reduction in sampling time compared to DiT \citep{peebles2022scalable}, while keeping the same NFE. Additionally, LEGO has the capability to generate images at much higher resolutions ($e.g.$, $2048\times 600$) than training images ($e.g.$, $256\times 256$).

\section{Local-feature Enrichment and Global-content Orchestration}\label{sec:method}

\subsection{Motivations for Constructing 
LEGO Bricks}

Both U-Net and ViT, which are two prominent architectural choices for diffusion models, impose a significant computational burden. The primary goal of this paper is to craft an efficient diffusion-modeling architecture that seamlessly integrates the principles of local feature enrichment, progressive spatial refinement, and iterative denoising within a unified framework. Additionally, our objective extends to a noteworthy capability: the selective skipping of LEGO bricks, which operate on different patch sizes at various time steps during the generation process. This unique test-time reconfigurable capability sets our design apart from prior approaches.

Our envisioned LEGO bricks are intended to possess several advantageous properties: \textbf{1) Spatial Efficiency in Training:} 
Within the ensemble, the majority of LEGO bricks are dedicated to producing local patches using computation-light MLP mixing and attention modules. This  design choice leads to a significant reduction in computational Floating-Point Operations (FLOPs) and substantially shortens the overall training duration.
\textbf{2) Spatial Efficiency in Sampling:} 
During sampling, the LEGO bricks can be selectively skipped at each time step without a discernible decline in generation performance. Specifically, when $t$ is large, indicating greater uncertainty in the global spatial structure, more patch-level LEGO bricks can be safely skipped. Conversely, when $t$ is small, signifying a more stable global spatial structure, more full-resolution LEGO bricks can be bypassed.
\textbf{3)
Versatility:} 
LEGO bricks showcase remarkable versatility, accommodating both end-to-end training and sequential training from lower to upper bricks, all while enabling generation at resolutions significantly higher than those employed during training. Furthermore, they readily support the integration of existing pre-trained models as LEGO bricks, enhancing the model's adaptability and ease of use.

\subsection{Technical Preliminary of Diffusion-based Generative Models}\label{sec:method-tech}
Diffusion-based generative models 
employ a forward diffusion chain to gradually corrupt the data into noise, and an iterative-refinement-based reverse diffusion chain to regenerate the data.
\rev{We use $\alpha_t\in [0,1]$, a decreasing sequence, to define the variance schedule $ \{\beta_t=1-\frac{\alpha_t}{\alpha_{t-1}}\}_{t=1}^T $ and denote $\rvx_0$ as a clean image and $\rvx_t$ as a noisy image}.
The reverse diffusion chain can be optimized by maximizing the evidence lower bound (ELBO) \citep{blei2017variational} of a variational autoencoder \citep{kingma2013auto}, using a hierarchical prior with $T$ stochastic layers \citep{ho2020denoising,song2020denoising,song2021scorebased,kingma2021variational,zhou2023beta}.
Let's represent the data distribution as $ q(\rvx_0) $ and the generative prior as $ p(\rvx_T) $. The forward and reverse diffusion processes discretized into $T$ steps can be expressed as:
\begin{align}
&\text{Forward}: \resizebox{.83\linewidth}{!}{$ q(\rvx_{0:T}) = q(\rvx_0) \prod_{t=1}^T q(\rvx_t\given \rvx_{t-1})=q(\rvx_0) \prod_{t=1}^T \mathcal N \left(\rvx_t; {\frac{\sqrt{\alpha_t}}{\sqrt{\alpha_{t-1}}}}\rvx_{t-1},1-\frac{\alpha_t}{\alpha_{t-1}}\right)$}, \\
&\text{Reverse}: \textstyle p_\rvtheta(\rvx_{0:T}) = p(\rvx_T) \prod_{t=1}^T p_\rvtheta(\rvx_{t-1}\given \rvx_t) = p(\rvx_T) \prod_{t=1}^Tq(\rvx_{t-1}\given \rvx_t,\hat{\rvx}_0=f_{\theta}(\rvx_ t,t)), \label{eq:generative_distribution}
\end{align}
where $\alpha_t\in [0,1]$ is a decreasing sequence, which determines the variance schedule $ \{\beta_t=1-\frac{\alpha_t}{\alpha_{t-1}}\}_{t=1}^T $, and
$
q(\rvx_{t-1}\given \rvx_{t},\rvx_{0}) 
$
is the conditional posterior of the forward diffusion chain that is also used to construct the reverse diffusion chain. 
A standout feature of this construction is that both the forward marginal and the reverse conditional are analytic, following the Gaussian distributions as
\begin{align}
q(\rvx_{t}\given\rvx_{0}) &= \mathcal{N}(\rvx_t; \sqrt{\alpha_t}\rvx_{0}, (1 - \alpha_t) \mathbf{I}), \label{eq:x_t_distribution} \\
q(\rvx_{t-1}\given \rvx_{t},\rvx_{0}) 
&=\textstyle \mathcal{N}
\big(\frac{\sqrt{\alpha_{t-1}}}{1-\alpha_{t}}(1-\frac{\alpha_t}{\alpha_{t-1}})\rvx_{0} + 
\frac{(1-\alpha_{t-1})\sqrt{{\alpha_t}}}{(1-\alpha_t)\sqrt{{\alpha_{t-1}} }}
\rvx_{t},~\frac{1-\alpha_{t-1}}{1-\alpha_t}(1-\frac{\alpha_t}{\alpha_{t-1}})\mathbf{I}
\big) .
\end{align}
Furthermore,  maximizing the ELBO can be conducted by sampling $t\in\{1,\ldots,T\}$ and  minimizing 
\begin{align}
&\textstyle L_t=\mathrm{KL}(q(\rvx_{t-1}\given \rvx_{t},\rvx_0)||q(\rvx_{t-1}\given \rvx_t,\hat{\rvx}_0=f_{\theta}(\rvx_ t,t))) = \frac{\text{SNR}_{t-1}-\text{SNR}_t}{2}\|\rvx_0-f_{\theta}(\rvx_t,t)\|_2^2, \label{eq:KL_loss}
\end{align}
where $\text{SNR}_t : = \alpha_t/(1-\alpha_t)$.
From \Eqref{eq:x_t_distribution}, we can sample $\rvx_t$ via reparameterization as $\rvx_t = \sqrt{\alpha_t}\rvx_{0}+\sqrt{1-\alpha_t}\rvepsilon,~ \rvepsilon \sim \mathcal{N}(\mathbf{0}, \mathbf{I})$, and hence we can parameterize $\hat{\rvx}_0$ as a linear combination of $\rvx_t$ and a noise prediction $\rvepsilon_{\theta}(\rvx_t,t)$ as
\begin{align}
\textstyle \hat{\rvx}_0(\rvx_t, t;\theta) = f_{\theta}(\rvx_t, t) = \frac{\rvx_t}{\sqrt{\alpha_t}} - \frac{\sqrt{1-\alpha_t}}{\sqrt{\alpha_t}} \rvepsilon _{\theta}(\rvx_t, t) . %
\label{eq:x0_hat_param}
\end{align}
Thus we can also express $L_t$ in \Eqref{eq:KL_loss} in terms of noise prediction as 
$L_t = \frac{1}{2}(\frac{\text{SNR}_{t-1}}{\text{SNR}_{t}}-1) \| \rvepsilon - \rvepsilon_{\theta}(\rvx_t,t)\|_2^2$.
In summary, the network parameter $\theta$ 
can be trained by predicting either $\rvx_0$~\citep{karras2022edm} or the injected noise $\rvepsilon$ in forward diffusion~\citep{ho2020denoising,song2021scorebased}:
\begin{align}
&\textstyle \rvtheta^\star = \argmin_\rvtheta \mathop{\mathbb{E}}_{t, \rvx_t, \rvepsilon} [\lambda_t^\prime \| \hat{\rvx}_0(\rvx_t,t;{\rvtheta}) - \rvx_0 \|_2^2] ,~\text{or}~ \rvtheta^\star = \argmin_\rvtheta \mathop{\mathbb{E}}_{t, \rvx_t, \rvepsilon} [\lambda_t \| \rvepsilon_{\rvtheta}(\rvx_t,t) - \rvepsilon \|_2^2], \label{eq:dpm training loss}
\end{align}
where $ \lambda_t,~\lambda_t^\prime $ are both time-dependent weight 
coefficients, which are often chosen to be different from the ones suggested by the ELBO to enhance image generation quality \citep{ho2020denoising}.

\subsection{Stackable and Skippable LEGO Bricks: Training and Sampling}

In our architectural design, we utilize the diffusion loss to train LEGO bricks at various levels of spatial granularity, with each level corresponding to specific spatial refinement steps. We employ Transformer blocks to capture global content at the patch level, aligning them with the specific LEGO brick in operation. 
Unlike U-Net, which convolves
local
filters across all pixel locations, our method enriches feature embeddings from non-overlapping local regions and does not include upsampling layers.
In contrast to ViT, our approach operates many of its network layers at local patches of various sizes. 
 This enables us to efficiently capture localized information and progressively aggregate it.

\begin{figure}[t]
  \centering\vspace{-8pt}
  \includegraphics[width=0.95\linewidth]{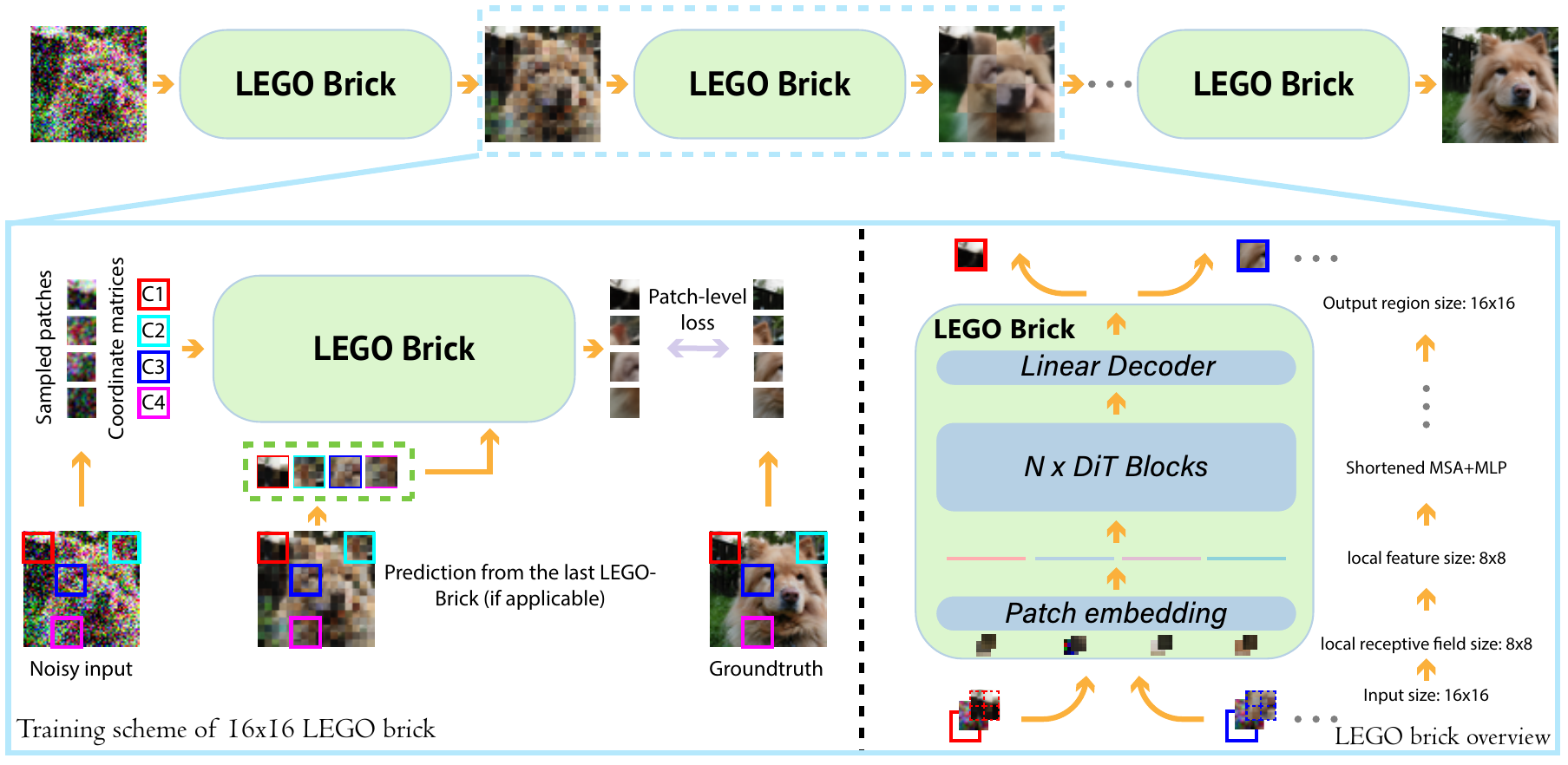}\vspace{-3mm}
  \caption{\small Top Panel: The generation paradigm with LEGO spatial refinement in a single time step. Bottom Panel:  An overview of the training scheme (\textit{Left}) and architecture (\textit{Right}) of a LEGO brick, {whose input size is $16\times 16$, local receptive field size is $8\times 8$, and attention span is 4}.}\vspace{-3mm}
  \label{fig:model}
\end{figure}

\subsubsection{LEGO Brick Ensemble and Vertical Spatial Refinement} %
\label{sec:spatial}

Diffusion models use their network backbone to iteratively refine image generation over the time dimension. When employing a backbone consisting of $K$ stacked LEGO bricks, we can progressively refine spatial features at each time step along the vertical dimension.
When a LEGO brick processes its patches, it independently refines each patch within the same brick. Additionally, we maintain the full-resolution image consistently at both the input and output of all bricks. This not only provides flexibility in choosing both patch sizes and locations for various bricks but also creates a test-time reconfigurable structure that can selectively skip bricks to significantly reduce sampling costs.

Denote the original image with spatial dimensions $H \times W$ as $\rvx$.
For the $k^{th}$ LEGO brick, which operates on patches of size $r_h(k)\times r_w(k)$, where $r_h(k)\le H$ and $r_w(k)\le  W$, we extract a set of patches of that size from $\rvx$. To simplify, we assume the brick size $r_h(k)=r_w(k)=r_{k}$, both $\frac{H}{r_{k}}$ and $\frac{W}{r_{k}}$ are integers, and the image is divided into non-overlapping patches, represented as: 
$$
\rvx^{(k)}_{(i,j)} \textstyle= \rvx[(i-1)r_{k}+1:i r_{k}, (j-1) r_{k}+1:jr_{k}]; ~i \in \{1, ..., \frac{H}{r_{k}}\}, ~j \in \{1, ..., \frac{W}{r_{k}}\}.
$$
We also denote $\rvm \in [-1,1]^{H\times W}$ as the normalized coordinates of the image pixels, and similarly, $\rvm_{(i,j)}^{(k)}$ as the coordinate matrix of the $(i,j)^{th}$ patch at the $k^{th}$ LEGO brick.

Mathematically, for the $k^{th}$ LEGO brick parameterized by $\theta_k$, and denoting $\rvz_{t,(i,j)}^{(k)}$ as the $(i,j)^{th}$ patch extracted at time $t$ from its output (with $\rvz_{t,(i,j)}^{(0)}=\emptyset$), the LEGO operation can be expressed as:
\ba{
\rvz_{t,(i,j)}^{(k)} = f_{\theta_k}(\rvx_{t,(i,j)}^{(k)}, \rvm_{(i,j)}^{(k)}, \rvz_{t,(i,j)}^{(k-1)}, t),
}
which receives patches at the same spatial locations from both the previous time step and the lower brick as its input.
For now, we treat this patch-level operation-based LEGO brick as a black box and will not delve into its details until we finish describing its recursive-based ensemble and training loss. 
To streamline the discussion, we will omit patch indices and coordinates when appropriate.

\textbf{Recursive ensemble of \rev{LEGO} bricks: }
The vanilla denoising diffusion step, as  in Equation \ref{eq:x0_hat_param}, is decomposed into $K$ consecutive LEGO bricks, stacked from the top to the bottom as follows:
\ba{
\hat{\rvx}_0(\rvx_t, t;\theta)=\rvz_t^{(K)},~\text{where}~\rvz_t^{(k)}=f_{\theta_k}(\rvx_t, \rvz^{(k-1)}_t, t)~\text{for}~k=K,\ldots,1, \label{eq:lego_module}
}
with $\rvz_t^{(0)}:=\emptyset$, $\theta:=\{\theta_k\}_{1,K}$, and
$\rvz^{(k)}_t$ denoting a grid of refined patches based on the corresponding patches from the output of the lower LEGO brick at time $t$.
We note that since each LEGO brick starts with a full-resolution image and ends with a refined full-resolution image, we have the flexibility to choose the target brick size for each LEGO brick. 
In our illustration shown in \Figref{fig:model}, we follow a progressive-growing-based construction, where the patch size of the stacked LEGO bricks monotonically increases when moving from lower to upper bricks. Alternatively, we can impose a monotonically decreasing constraint, corresponding to progressive refinement. 
In this work, we explore both progressive growth and refinement as two special examples, %
and further present a Unet-inspired structural variant that combines both. %
The combinatorial optimization of brick sizes for the LEGO bricks in the stack is a topic that warrants further investigation.

\textbf{Training loss: }
Denoting a noise-corrupted patch at time $t$ as $\rvx_t^{(k)}$, we have the diffusion chains as 
\begin{align}
&\text{Forward}: \textstyle q(\rvx^{(k)}_{0:T}) 
=q(\rvx^{(k)}_0) \prod_{t=1}^T \mathcal N \left(\rvx_t^{(k)}; {\frac{\sqrt{\alpha_t}}{\sqrt{\alpha_{t-1}}}}\rvx_{t-1}^{(k)},1-\frac{\alpha_t}{\alpha_{t-1}}\right), \\
&\text{Reverse}: \textstyle p_\rvtheta(\rvx_{0:T}) 
= p(\rvx_T) \prod_{t=1}^Tq(\rvx_{t-1}^{(k)}\given \rvx_t^{(k)},\hat{\rvx}_0^{(k)}=f_{\theta}(\rvx^{(k)}_ t,\rvx_t^{(k-1)},t)), \label{eq:generative_distribution_lego}
\end{align}
Denote $\epsilon$ as a normal noise used to corrupt the clean image. The clean image patches, denoted as $\rvx_{0,(i,j)}^{(k)}$, are of size $r_{k}\times r_{k}$ and correspond to specific spatial locations $(i,j)$. Additionally, we have refined patches, $\hat{\rvx}_{0,(i,j)}^{(k)}$, produced by the $k^{th}$ LEGO brick at the same locations, and ${\hat{\rvx}}_{0,(i,j)}^{(k-1)}$ from the $(k-1)^{th}$ LEGO brick.
When processing a noisy image $\rvx_t$ at time $t$, we perform upward propagation through the stacked LEGO bricks to progressively refine the full image. This begins with $\hat{\rvx}_0^{(0)}=\emptyset$ and proceeds with refined image patches $\hat{\rvx}_0^{(k)}$ for $k=1,\ldots,K$. As each LEGO brick operates on sampled patches, when the number of patches is limited, there may be instances where $\hat{\rvx}_{0,(i,j)}^{(k-1)}$ contain missing values. In such cases, we replace these missing values with the corresponding pixels from $\rvx_0$ to ensure the needed $\hat{\rvx}_{0,(i,j)}^{(k-1)}$ is present.
Denote
$\lambda_t^{(k)}$ as time- and brick-dependent weight coefficients, whose settings are described in Appendix~\ref{sec:experiment-detail}. With the refined image patches $\hat{\rvx}_{0,(i,j)}^{(k)}$, 
we express the training loss %
over the $K$ LEGO bricks as
\ba{
\textstyle \mathop{\E_k\mathbb{E}}_{t, \rvx_0^{(k)}, \rvepsilon, (i,j)} [\lambda_t^{(k)} \|  \rvx_{0,(i,j)}^{(k)} -\hat{\rvx}_{0,(i,j)}^{(k)}\|_2^2],~~\hat{\rvx}_{0,(i,j)}^{(k)}:=f_{\theta_k}(\rvx^{(k)}_ {t,(i,j)}, {\hat{\rvx}}_{0,(i,j)}^{(k-1)}, %
t).
}
This training scheme enables each LEGO brick to refine local patches while being guided by the noise-corrupted input and the output of the previous brick in the stack.

\subsubsection{Network Design within the LEGO Brick}\label{sec:local-global}

In the design of a LEGO brick, we draw inspiration from the design of the Diffusion Transformer (DiT) proposed in \citet{peebles2022scalable}. The DiT architecture comprises several key components:
Patch Embedding Layer: Initially, the input image is tokenized into a sequence of patch embeddings;
DiT Blocks: These are Transformer blocks featuring multi-head attention, zero-initialized adaptive layer norm (adaLN-zero), and MLP layers;
\rev{Linear layer with adaLN: Positioned on top of the DiT blocks, this linear layer, along with adaLN, converts the sequence of embeddings back into an image.}
The diffusion-specific conditions, such as time embedding and class conditions, are encoded within the embeddings through adaLN in every block.

As illustrated in \Figref{fig:model}, we employ the DiT blocks, along with the patch-embedding layer and the linear decoder to construct a LEGO brick. The input channel dimension of the patch-embedding layer is expanded to accommodate the channel-wise concatenated input $[\rvx_t^{(k)}, \hat{\rvx}_0^{(k-1)}, \rvm^{(k)}]$. 
{
The size of the local receptive fields, denoted as $\ell_{k}\times \ell_{k}$, %
is subject to variation to achieve spatial refinement. In this paper, by default, we set it as follows: if the brick size $r_{k}\times r_{k}$ is smaller than the image resolution, it is set to $r_{k}/2\times r_{k}/2$, and if the brick size is equal to the image resolution, it is set to $2\times 2$. In simpler terms, for the $k^{th}$ LEGO brick, when $r_{k}\times r_{k}$ is smaller than $W\times H$ (or $W/8\times H/8$ for diffusion in the latent space), we partition each of the input patches of size $r_{k}\times r_{k}$ into four non-overlapping local receptive fields of size $r_{k}/2\times r_{k}/2$. These four local receptive fields are further projected by the same MLP to become a sequence of four token embeddings. Afterward, these four token embeddings are processed by the DiT blocks, which have an attention span as short as four, and decoded to produce output patches of size $r_{k}\times r_{k}$.}

{We categorize the LEGO bricks as `patch-bricks' if their brick size is smaller than the image resolution, and as `image-bricks' otherwise. The patch-bricks have attention spans as short as four, saving memory and computation in both training and generation. While the image-bricks choose longer attention spans and hence do not yield memory or computation savings during training, they can be compressed with fewer DiT blocks compared with the configuration in \citet{peebles2022scalable}. During generation, LEGO bricks can also be selectively skipped at appropriate time steps without causing clear performance degradation, resulting in substantial savings in generation costs.}
For simplicity, we maintain a fixed embedding size and vary only the number of DiT blocks in each brick based on $r_{k}$. Each LEGO brick may require a different capacity depending on $r_{k}$. For a comprehensive configuration of the LEGO bricks, please refer to Table \ref{tab:model_config} in Appendix~\ref{sec:experiment-detail}.

We explore three spatial refinement settings and stack LEGO bricks accordingly: 1) Progressive Growth (PG): In this setup, we arrange LEGO bricks in a manner where each subsequent brick has a patch size that's four times as large as the previous one, $i.e.$, $r_{k} = 4r_{k-1}$. Consequently, the patch of the current break would aggregate the features of four patches output by the previous brick, facilitating global-content orchestration. 2)
Progressive Refinement (PR): By contrast, for the PR configuration, we stack LEGO bricks in reverse order compared to PG, with $r_{k} = r_{k-1}/4$. Here, each LEGO brick is responsible for producing a refined generation based on the content provided by the brick below it. 
{3) LEGO-U: a hierarchical variant that combines the features of LEGO-PR and LEGO-PG. This model processes image patches starting with larger resolutions, transitioning to smaller ones, and then reverting back to larger resolutions, similar to Unets that operate with multiple downsampling and upsampling stages to have multi-scale representations. }

\section{Experiments}\label{sec:experiment}

We present a series of experiments designed to assess the efficacy and versatility of using stacked LEGO bricks as the backbone for diffusion models. We defer the details on datasets, training settings, and evaluation metrics into Appendix~\ref{sec:experiment-detail}. We begin by showcasing results on both small-scale and large-scale datasets to evaluate the image generation performance using the three configurations outlined in Table \ref{tab:model_config} in Appendix~\ref{sec:experiment-detail}.

\subsection{Main results}
\textbf{Training in pixel space: } 
We begin by comparing LEGO in pixel-space training on CelebA and ImageNet with two Transformer-based state-of-the-art diffusion models \citep{bao2022all,peebles2022scalable}, which are our most relevant baselines. All images are resized to a resolution of $64\times64$. We conduct experiments with two spatial refinement methods using LEGO, denoted as -PG for the progressive growth variant and -PR for the progressive refinement variant, as explained in Section~\ref{sec:local-global}. 
In both Tables~\ref{tab:CelebA64} and~\ref{tab:IN64}, when compared to DiT, LEGO slightly increases the number of parameters due to the expansion of the input dimension in a channel-wise manner. However, in terms of FLOPs, we observe that LEGO incurs a significantly lower computational cost compared to both U-ViT and DiT. Despite maintaining such a low computational cost, LEGO still wins the competition in terms of FID performance.

\begin{table}[t]
\begin{minipage}{0.48\textwidth}
\caption{FID results of unconditional image generation on CelebA $64\times 64$~\citep{celebA2015deep}. All baselines use a Transformer-based backbone.}\vspace{-2mm}\label{tab:CelebA64}
\setlength{\tabcolsep}{1.0mm}{ 
\scalebox{0.94}{
\begin{tabular}{lccc}
\toprule[1.5pt]
Method & \#Params & FLOPs & FID \\
\midrule
U-ViT-S/4 & 44M &  1.2G  &  2.87 \\
DiT-S/2 & 33M &  0.4G &   {2.52} \\
LEGO-S-PG (ours) & 35M & 0.2G  &   {2.17} \\
LEGO-S-PR (ours) & 35M & 0.2G  &   \textbf{2.09} \\
\bottomrule[1.5pt]
\end{tabular}
}}
\end{minipage}
\hfill
\begin{minipage}{0.48\textwidth}
\caption{FID results of conditional image generation on ImageNet $64\times 64$~\citep{deng2009imagenet}. All baselines use a Transformer-based backbone.}\vspace{-2mm}
\label{tab:IN64}
\setlength{\tabcolsep}{1.0mm}{ 
\scalebox{0.94}{
\begin{tabular}{lccc}
\toprule[1.5pt]
Method & \#Params & FLOPs & FID \\
\midrule
U-ViT-L/4 & 287M &  91.2G  &  4.26 \\
DiT-L/2 & 458M & 161.4G  &   {2.91} \\
LEGO-L-PG (ours) & 464M & 68.8G  &  \textbf{2.16} \\
LEGO-L-PR (ours) & 464M & 68.8G  &   2.29  \\
\bottomrule[1.5pt]
\end{tabular}
}}
\end{minipage}\vspace{-3mm}
\end{table}

\begin{table}[t]
\centering
\small
\caption{\small Comparison of LEGO-diffusion with state-of-the-art generative models on class-conditional generation using ImageNet at resolutions 256$\times$256 and $512 \times 512$. Each metric is presented in two columns, one without classifier-free guidance and one with, marked with \xmark~ and \cmark, respectively.}\vspace{-2mm}\label{tab:IN256}
\centering
\setlength{\tabcolsep}{1.0mm}{ 
\scalebox{0.93}{
\centering
\begin{tabular}{lcc|cc|cc|cc|cc|c}
\toprule[1.5pt]
Evaluation~Metric      & \multicolumn{2}{c}{FID$\downarrow$}   & \multicolumn{2}{c}{sFID$\downarrow$} & \multicolumn{2}{c}{IS$\uparrow$}     & \multicolumn{2}{c}{Prec.$\uparrow$} & \multicolumn{2}{c}{Rec.$\uparrow$}  & {Iterated}\\ \cline{1-12}
Classifier-free guidance 
& \xmark & \cmark  & \xmark & \cmark  & \xmark & \cmark  & \xmark & \cmark  & \xmark & \cmark & Images (M)\\
\midrule
\multicolumn{11}{l}{\textcolor{gray}{U-Net-based architecture ($256 \times 256$)}} \\
ADM~\citep{dhariwal2021diffusion}     & 10.94 & 4.59 & 6.02 & 5.25 & 100.98 & 186.70 & 0.69 & 0.82 & 0.63 & 0.52 & \multirow{2}{*}{506}\\
ADM-Upsampled    & 7.49 & 3.94 & {5.13}  & 6.14 & 127.49  & 215.84 & 0.72 & 0.83 & 0.63 & 0.53  &   \\
LDM-8~\citep{rombach2022high}    & 15.51 & 7.76 & -  & -    & 79.03 & 103.49 & 0.65 & 0.71 & 0.63 & \textbf{0.62} &  307  \\
LDM-4    & 10.56 & 3.60  & -  & -    & 103.49  & 247.67  & 0.71 & \textbf{0.87} & 0.62 & 0.48  & 213 \\ 
\arrayrulecolor{gray}\midrule
\multicolumn{12}{l}{\textcolor{gray}{Transformer-based architecture ($256 \times 256$)}} \\ 
U-ViT-H/2~\citep{bao2022all}  & 6.58 & 2.29 & - & 5.68 & - & 263.88 & - & 0.82 & - & 0.57 & 307\\
DiT-XL/2~\citep{peebles2022scalable} & 9.62  & 2.27 & 6.85  & 4.60 & 121.50 & 278.24 & 0.67 & 0.83 & \textbf{0.67} & 0.57 &  1792\\
MDT-XL/2~\citep{gao2023masked}     & 6.23 & \textbf{1.79}  & 5.23 & {4.57} & 143.02 & {283.01} & 0.71 & 0.81 & 0.65 & 0.6 & 1664\\
{MaskDiT}~\citep{zheng2023fast}     & {5.69}  & 2.28  & 10.34 & 5.67     & 177.99    & 276.56        & 0.74    & 0.80     & 0.60  &  0.61 & 521\\ 
\arrayrulecolor{gray}\midrule
{LEGO}-XL-PG (ours, $256 \times 256$)      & {7.15} & {2.05}   & 7.71 & 4.77     & 192.75    & {{289.12}}      &\textbf{ 0.77}      & 0.84   & 0.63    & 0.55 &  512\\
{LEGO}-XL-PR (ours, $256 \times 256$)     & {5.38} & {2.35}   & 9.06 & 5.21     & 189.12    & {284.73}        & \textbf{0.77}      & 0.83   & 0.63    & 0.60 &  512 \\
{LEGO}-XL-U (ours, $256 \times 256$)     & \textbf{5.28} & {2.59}   & \textbf{4.91} &\textbf{4.21}     & \textbf{337.85}    & \textbf{338.08}        & {0.67}      & 0.83   & 0.65    & 0.56 &  350 \\
\arrayrulecolor{black}\bottomrule
\toprule
\multicolumn{12}{l}{\textcolor{gray}{U-Net-based architecture ($512 \times 512$)}} \\
ADM~\citep{dhariwal2021diffusion}      & 23.24 & 7.72 & 10.19 & 6.57 & 58.06 & 172.71 & 0.73 & \textbf{0.87} & 0.60 & 0.42  &  \multirow{2}{*}{496}\\
ADM-Upsampled     & 9.96 & 3.85 & \textbf{5.62} & 5.86 & 121.78 & 221.72 & 0.75 & 0.84 & \textbf{0.64} & 0.53  &   \\
\arrayrulecolor{gray}\midrule
\multicolumn{12}{l}{\textcolor{gray}{Transformer-based architecture ($512 \times 512$)}} \\ 
DiT-XL/2~\citep{peebles2022scalable} & 12.03  & \textbf{3.04} & 7.12  & 5.02 & 105.25 & 240.82 & 0.75 & 0.84 & \textbf{0.64} & 0.54  & 768  \\
\arrayrulecolor{gray}\midrule
{LEGO}-XL-PG (ours, $512 \times 512$)     & 10.06 & 3.74   & 5.96 & \textbf{4.62}     & \textbf{192.27}    & \textbf{285.66}      & \textbf{0.77}      & 0.85   & \textbf{0.64 }   & \textbf{0.64} &  512 \\
{LEGO}-XL-PR (ours, $512 \times 512$)    & \textbf{9.01} & \rev{3.99}   &  6.39 & \rev{4.87}     & 168.92    & \rev{265.75}        & 0.76      & \rev{0.86}   & 0.63    & \rev{0.49}  &   512 \\
\arrayrulecolor{black}\bottomrule[1.5pt]
\end{tabular}
}}\vspace{-6mm}
\end{table}

\textbf{Training in latent space: } 
We conduct comparisons with state-of-the-art class-conditional generative models trained on ImageNet at resolutions of $256 \times 256$ and $512 \times 512$, as shown in Table~\ref{tab:IN256}. Our results are obtained after training with 512M images (calculated as the number of iterations $\times$ batch size). When considering LEGO without classifier-free guidance (CFG), we observe that it achieves a better FID than the baseline diffusion models, striking a good balance between precision and recall. One potential explanation for this improvement is that the LEGO bricks with small patch sizes enhance patch diversity, preserving both generation fidelity and diversity.

With the inclusion of CFG, we notably achieve an FID of 2.05 and the IS of 289.12 on ImageNet-$256 \times 256$. It is notable that the LEGO-U variant, which incorporates the features of the -PR and -PG variants, has achieved a clearly higher IS (338.08 \textit{w.} guidance and 337.85 \textit{w.o.} guidance). This result significantly surpasses the baselines and demonstrates the power of multi-scale information in generative modeling. In the case of using $512 \times 512$ resolution data, LEGO also achieves the best IS and produces a competitive FID with fewer iterated images. Additionally, as depicted in \Figref{fig:conv_flops}, LEGO demonstrates superior convergence speed and requires significantly less training time. For qualitative validation, we provide visualizations of randomly generated samples in \Figref{fig:imagenet-vis}.

\subsection{\rev{Improving the efficiency of diffusion modeling with LEGO}}\label{sec:skip_sampling}
The design of LEGO inherently facilitates the sampling process by generating images with selected LEGO bricks. Intuitively, at low-noise timesteps ($i.e.$, when $t$ is small), the global structure of images is {already} well-defined, and {hence} the model {is desired to} prioritize {local details}. Conversely, when images are noisier, the global-content orchestration becomes crucial to {uncover global structure under high uncertainties.} 
Therefore, for improved efficiency, we skip LEGO bricks that emphasize local details during high-noise timesteps and those that construct global structures during low-noise timesteps.
To validate this design, we conduct experiments to study the generation performance when LEGO bricks are skipped {in different degrees}. 

{For LEGO-PG, we designate a timestep $t_\text{break}$ as a breakpoint. When $t > t_\text{break}$, no bricks are skipped, but when $t \leq t_\text{break}$, the top-level brick, which is the image-brick with a brick size of $64\times 64$, is skipped for the remainder of the sampling process. In contrast, for PR, we set a timestep $t_\text{break}$ as a breakpoint. When $t \le T - t_\text{break}$, no bricks are skipped, but when $t > T - t_\text{break}$, the top-level brick, which is the patch-break with a size of $4\times 4$, is skipped for the remainder of the sampling process. 

We evaluate the impact of the choice of $t_\text{break}$ on both FID scores and sampling time for LEGO-PG and LEGO-PR. \rev{The results in \Figref{fig:sampling_skip_IN256} illustrate the FID scores and the time required to generate 50,000 images using 8 NVIDIA A100 GPUs}. In LEGO-PG, when we skip the top-level LEGO brick, we observe a notable enhancement in sampling efficiency, primarily because this brick, responsible for processing the entire image, is computationally demanding. Nevertheless, skipping this brick during timesteps with high levels of noise also results in a trade-off with performance, as the model loses access to global content. On the other hand, for LEGO-PR, skipping the top-level patch-brick slightly impacts performance when $t_\text{break}$ is large, while the improvement in sampling efficiency is not as substantial as with LEGO-PG. Interestingly, when $t_\text{break}$ is chosen to be close to the halfway point of sampling, performance is preserved for both models, and significant time savings can be achieved during sampling.

\begin{figure}[t]
\centering
    \includegraphics[width=\textwidth]{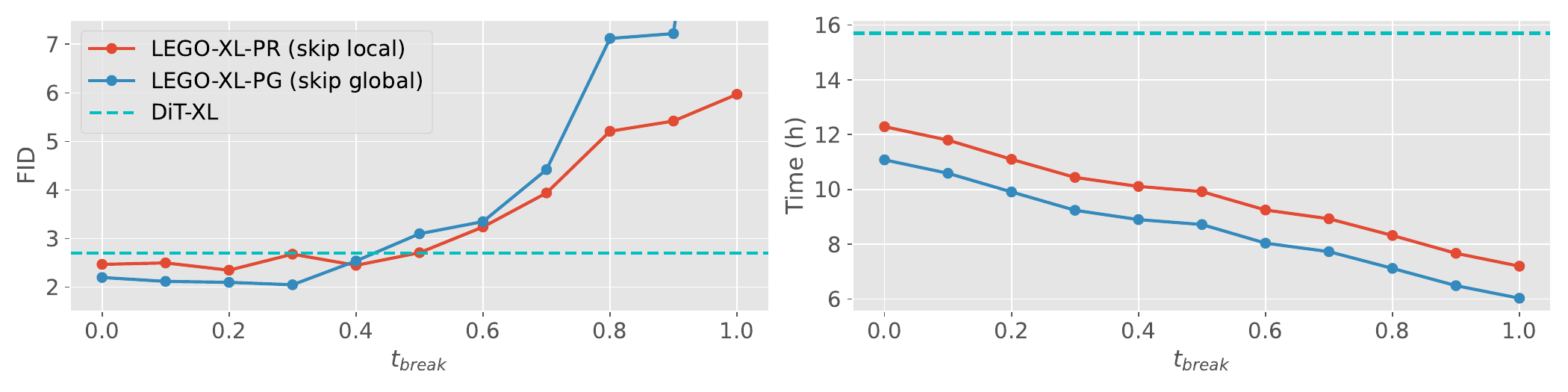} \vspace{-8mm}
    \caption{\rev{\small Visualization of how FID and inference time for 50k images change as the proportion of reverse diffusion time steps, at which the top-level brick is skipped, increases from 0 to 1. LEGO-PG prioritizes skipping the top-level brick at earlier time steps, while LEGO-PR prioritizes it at later time steps. The experiments are conducted using the LEGO-XL model with eight NVIDIA A100 GPUs on the ImageNet ($256\times 256$) dataset.}}\label{fig:sampling_skip_IN256}  \vspace{-6mm}
\end{figure}

\subsection{\rev{Leveraging a pretrained diffusion model as a LEGO brick}}\label{sec:a_LEBO_brick}
Apart from the training efficiency, another advantage of the design of LEGO is any pre-trained diffusion models can be incorporated as a LEGO brick as long as they can predict $\hat{\rvx}_0^{(k)}$. A natural choice is to deploy a pre-trained diffusion model as the first LEGO brick in LEGO-PR. Then, a series of LEGO bricks that refine local details can be built on top of it. To validate this concept, we conducted experiments on CIFAR-10 using LEGO-PR. We deploy an unconditional Variance Preservation (VP) diffusion model~\citep{song2021scorebased} (implemented under the EDM codebase provided by \citet{karras2022edm}) to replace the LEGO brick that deals with full images, and build a LEGO-S ($4 \times 4$) brick on the top for refinement. We train the incorporated model in two different ways:  1) VP (frozen) + LEGO-PR: The VP model is fixed and only the other LEGO bricks are trained, and 2) VP (unfrozen) + LEGO-PR: The VP model is fine-tuned along with the training of the other LEGO bricks. Table~\ref{tab:versatile} shows that the combined model consistently improves the base model to obtain a better generation result, demonstrating the versatility of LEGO bricks.

\begin{table}[h]
  \centering
  \caption{\rev{
  Outcomes from CIFAR10 on integrating a pre-trained model as the first brick in LEGO-PR.
  }} \label{tab:versatile} \vspace{-6pt}
\setlength{\tabcolsep}{1.0mm}{ 
\scalebox{1.0}{
  \begin{tabular}{l|ccc}
    \toprule[1.5pt]
    Model &{\text{~~~~~}  VP\text{~~~~~}    }& VP (frozen) + LEGO-PR & \text{~~}VP (unfrozen) + LEGO-PR\text{~~} \\
    \midrule
    FID & 2.5&   1.92 &  1.88   \\
    \bottomrule[1.5pt]
  \end{tabular}
}}
\end{table}

\section{Conclusion}\label{sec:conclusion}

We introduce the LEGO brick, a novel network unit that seamlessly integrates Local-feature Enrichment and Global-content Orchestration. These bricks can be stacked to form the backbone of diffusion models. LEGO bricks are adaptable, operating on local patches of varying sizes while maintaining the full-resolution image across all bricks. During training, they can be flexibly stacked to construct diffusion backbones of diverse capacity, and during testing, they are skippable, enabling efficient image generation with a smaller network than used in training. This design choice not only endows LEGO with exceptional efficiency during both training and sampling but also empowers it to generate images at resolutions significantly higher than training. Our extensive experiments, conducted on challenging image benchmarks such as CelebA and ImageNet, provide strong evidence of LEGO's ability to strike a compelling balance between computational efficiency and generation quality. LEGO accelerates training, expedites convergence, and consistently maintains, if not enhances, the quality of generated images. This work signifies a noteworthy advancement in addressing the challenges associated with diffusion models, making them more versatile and efficient for generating high-resolution photo-realistic images.

\subsubsection*{Acknowledgments}
H. Zheng, Z. Wang, and M. Zhou acknowledge the support of NSF-IIS 2212418, NIH-R37 CA271186, and the NSF AI Institute for
Foundations of Machine Learning (IFML).

\bibliography{iclr2024_conference}

\begin{thebibliography}{83}
\providecommand{\natexlab}[1]{#1}
\providecommand{\url}[1]{\texttt{#1}}
\expandafter\ifx\csname urlstyle\endcsname\relax
  \providecommand{\doi}[1]{doi: #1}\else
  \providecommand{\doi}{doi: \begingroup \urlstyle{rm}\Url}\fi

\bibitem[Aharon et~al.(2006)Aharon, Elad, and Bruckstein]{aharon2006k}
Michal Aharon, Michael Elad, and Alfred Bruckstein.
\newblock {K-SVD}: An algorithm for designing overcomplete dictionaries for
  sparse representation.
\newblock \emph{IEEE Transactions on signal processing}, 54\penalty0
  (11):\penalty0 4311--4322, 2006.

\bibitem[Arakawa et~al.(2023)Arakawa, Tsunashima, Horita, Tanaka, and
  Morishima]{arakawa2023memory}
Shinei Arakawa, Hideki Tsunashima, Daichi Horita, Keitaro Tanaka, and Shigeo
  Morishima.
\newblock Memory efficient diffusion probabilistic models via patch-based
  generation.
\newblock \emph{arXiv preprint arXiv:2304.07087}, 2023.

\bibitem[Banerjee et~al.(2005)Banerjee, Merugu, Dhillon, Ghosh, and
  Lafferty]{banerjee2005clustering}
Arindam Banerjee, Srujana Merugu, Inderjit~S Dhillon, Joydeep Ghosh, and John
  Lafferty.
\newblock Clustering with {B}regman divergences.
\newblock \emph{Journal of machine learning research}, 6\penalty0 (10), 2005.

\bibitem[Bao et~al.(2022)Bao, Li, Cao, and Zhu]{bao2022all}
Fan Bao, Chongxuan Li, Yue Cao, and Jun Zhu.
\newblock All are worth words: A {ViT} backbone for score-based diffusion
  models.
\newblock \emph{arXiv preprint arXiv:2209.12152}, 2022.

\bibitem[Bao et~al.(2023)Bao, Nie, Xue, Li, Pu, Wang, Yue, Cao, Su, and
  Zhu]{bao2023one}
Fan Bao, Shen Nie, Kaiwen Xue, Chongxuan Li, Shi Pu, Yaole Wang, Gang Yue, Yue
  Cao, Hang Su, and Jun Zhu.
\newblock One transformer fits all distributions in multi-modal diffusion at
  scale.
\newblock \emph{arXiv preprint arXiv:2303.06555}, 2023.

\bibitem[Bar-Tal et~al.(2023)Bar-Tal, Yariv, Lipman, and
  Dekel]{bar2023multidiffusion}
Omer Bar-Tal, Lior Yariv, Yaron Lipman, and Tali Dekel.
\newblock Multidiffusion: Fusing diffusion paths for controlled image
  generation.
\newblock \emph{arXiv preprint arXiv:2302.08113}, 2023.

\bibitem[Bengio \& Bengio(1999)Bengio and Bengio]{bengio1999modeling}
Yoshua Bengio and Samy Bengio.
\newblock Modeling high-dimensional discrete data with multi-layer neural
  networks.
\newblock \emph{Advances in Neural Information Processing Systems}, 12, 1999.

\bibitem[Blei et~al.(2017)Blei, Kucukelbir, and McAuliffe]{blei2017variational}
David~M Blei, Alp Kucukelbir, and Jon~D McAuliffe.
\newblock Variational inference: A review for statisticians.
\newblock \emph{Journal of the American statistical Association}, 112\penalty0
  (518):\penalty0 859--877, 2017.

\bibitem[Choi et~al.(2021)Choi, Kim, Jeong, Gwon, and Yoon]{choi2021ilvr}
Jooyoung Choi, Sungwon Kim, Yonghyun Jeong, Youngjune Gwon, and Sungroh Yoon.
\newblock Ilvr: Conditioning method for denoising diffusion probabilistic
  models.
\newblock \emph{arXiv preprint arXiv:2108.02938}, 2021.

\bibitem[Chung et~al.(2022)Chung, Sim, Ryu, and Ye]{chung2022improving}
Hyungjin Chung, Byeongsu Sim, Dohoon Ryu, and Jong~Chul Ye.
\newblock Improving diffusion models for inverse problems using manifold
  constraints.
\newblock In Alice~H. Oh, Alekh Agarwal, Danielle Belgrave, and Kyunghyun Cho
  (eds.), \emph{Advances in Neural Information Processing Systems}, 2022.
\newblock URL \url{https://openreview.net/forum?id=nJJjv0JDJju}.

\bibitem[Deng et~al.(2009)Deng, Dong, Socher, Li, Li, and
  Fei-Fei]{deng2009imagenet}
Jia Deng, Wei Dong, Richard Socher, Li-Jia Li, Kai Li, and Li~Fei-Fei.
\newblock {ImageNet}: A large-scale hierarchical image database.
\newblock In \emph{2009 IEEE conference on computer vision and pattern
  recognition}, pp.\  248--255. Ieee, 2009.

\bibitem[Dhariwal \& Nichol(2021)Dhariwal and Nichol]{dhariwal2021diffusion}
Prafulla Dhariwal and Alex Nichol.
\newblock Diffusion models beat {GAN}s on image synthesis.
\newblock \emph{Advances in Neural Information Processing Systems (NeurIPS)},
  2021.

\bibitem[Ding et~al.(2022)Ding, Wang, Xu, Welch, and Wang]{ding2022continuous}
Xin Ding, Yongwei Wang, Zuheng Xu, William~J Welch, and Z~Jane Wang.
\newblock Continuous conditional generative adversarial networks: Novel
  empirical losses and label input mechanisms.
\newblock \emph{IEEE Transactions on Pattern Analysis and Machine
  Intelligence}, 2022.

\bibitem[Ding et~al.(2023)Ding, Zhang, Wu, and Tu]{ding2023patched}
Zheng Ding, Mengqi Zhang, Jiajun Wu, and Zhuowen Tu.
\newblock Patched denoising diffusion models for high-resolution image
  synthesis.
\newblock \emph{arXiv preprint arXiv:2308.01316}, 2023.

\bibitem[Dinh et~al.(2015)Dinh, Krueger, and Bengio]{dinh2014nice}
Laurent Dinh, David Krueger, and Yoshua Bengio.
\newblock {NICE}: Non-linear independent components estimation.
\newblock \emph{International Conference in Learning Representations Workshop
  Track}, 2015.

\bibitem[Dinh et~al.(2017)Dinh, Sohl{-}Dickstein, and Bengio]{dinh2016density}
Laurent Dinh, Jascha Sohl{-}Dickstein, and Samy Bengio.
\newblock Density estimation using real {NVP}.
\newblock In \emph{5th International Conference on Learning Representations,
  {ICLR} 2017, Toulon, France, April 24-26, 2017, Conference Track
  Proceedings}. OpenReview.net, 2017.
\newblock URL \url{https://openreview.net/forum?id=HkpbnH9lx}.

\bibitem[Dosovitskiy et~al.(2021)Dosovitskiy, Beyer, Kolesnikov, Weissenborn,
  Zhai, Unterthiner, Dehghani, Minderer, Heigold, Gelly, Uszkoreit, and
  Houlsby]{dosovitskiy2021an}
Alexey Dosovitskiy, Lucas Beyer, Alexander Kolesnikov, Dirk Weissenborn,
  Xiaohua Zhai, Thomas Unterthiner, Mostafa Dehghani, Matthias Minderer, Georg
  Heigold, Sylvain Gelly, Jakob Uszkoreit, and Neil Houlsby.
\newblock An image is worth 16x16 words: Transformers for image recognition at
  scale.
\newblock In \emph{International Conference on Learning Representations}, 2021.
\newblock URL \url{https://openreview.net/forum?id=YicbFdNTTy}.

\bibitem[Efron(2011)]{efron2011tweedie}
Bradley Efron.
\newblock Tweedie’s formula and selection bias.
\newblock \emph{Journal of the American Statistical Association}, 106\penalty0
  (496):\penalty0 1602--1614, 2011.

\bibitem[Esser et~al.(2021)Esser, Rombach, and Ommer]{esser2021taming}
Patrick Esser, Robin Rombach, and Bjorn Ommer.
\newblock Taming transformers for high-resolution image synthesis.
\newblock In \emph{Proceedings of the IEEE/CVF conference on computer vision
  and pattern recognition}, pp.\  12873--12883, 2021.

\bibitem[Gao et~al.(2023)Gao, Zhou, Cheng, and Yan]{gao2023masked}
Shanghua Gao, Pan Zhou, Ming-Ming Cheng, and Shuicheng Yan.
\newblock Masked diffusion transformer is a strong image synthesizer.
\newblock \emph{arXiv preprint arXiv:2303.14389}, 2023.

\bibitem[Goodfellow et~al.(2014)Goodfellow, Pouget-Abadie, Mirza, Xu,
  Warde-Farley, Ozair, Courville, and Bengio]{goodfellow2014generative}
Ian Goodfellow, Jean Pouget-Abadie, Mehdi Mirza, Bing Xu, David Warde-Farley,
  Sherjil Ozair, Aaron Courville, and Yoshua Bengio.
\newblock Generative adversarial nets.
\newblock In \emph{Advances in neural information processing systems}, pp.\
  2672--2680, 2014.

\bibitem[Gu et~al.(2022)Gu, Chen, Bao, Wen, Zhang, Chen, Yuan, and
  Guo]{gu2021vector}
Shuyang Gu, Dong Chen, Jianmin Bao, Fang Wen, Bo~Zhang, Dongdong Chen, Lu~Yuan,
  and Baining Guo.
\newblock Vector quantized diffusion model for text-to-image synthesis.
\newblock In \emph{CVPR}, 2022.

\bibitem[Ho et~al.(2020)Ho, Jain, and Abbeel]{ho2020denoising}
Jonathan Ho, Ajay Jain, and Pieter Abbeel.
\newblock Denoising {D}iffusion {P}robabilistic {M}odels.
\newblock \emph{Advances in Neural Information Processing Systems}, 33, 2020.

\bibitem[Ho et~al.(2022)Ho, Saharia, Chan, Fleet, Norouzi, and
  Salimans]{ho2022cascaded}
Jonathan Ho, Chitwan Saharia, William Chan, David~J Fleet, Mohammad Norouzi,
  and Tim Salimans.
\newblock Cascaded diffusion models for high fidelity image generation.
\newblock \emph{J. Mach. Learn. Res.}, 23\penalty0 (47):\penalty0 1--33, 2022.

\bibitem[Karras et~al.(2022)Karras, Aittala, Aila, and Laine]{karras2022edm}
Tero Karras, Miika Aittala, Timo Aila, and Samuli Laine.
\newblock Elucidating the design space of diffusion-based generative models.
\newblock In \emph{Proc. NeurIPS}, 2022.

\bibitem[Kim \& Ye(2021)Kim and Ye]{kim2021diffusionclip}
Gwanghyun Kim and Jong~Chul Ye.
\newblock {DiffusionCLIP}: Text-guided image manipulation using diffusion
  models.
\newblock \emph{arXiv preprint arXiv:2110.02711}, 2021.

\bibitem[Kingma \& Welling(2014)Kingma and Welling]{kingma2013auto}
Diederik~P. Kingma and Max Welling.
\newblock Auto-encoding variational {B}ayes.
\newblock In \emph{International Conference on Learning Representations}, 2014.

\bibitem[Kingma et~al.(2021)Kingma, Salimans, Poole, and
  Ho]{kingma2021variational}
Diederik~P Kingma, Tim Salimans, Ben Poole, and Jonathan Ho.
\newblock Variational diffusion models.
\newblock \emph{arXiv preprint arXiv:2107.00630}, 2021.

\bibitem[Kingma \& Dhariwal(2018)Kingma and Dhariwal]{kingma2018glow}
Durk~P Kingma and Prafulla Dhariwal.
\newblock Glow: Generative flow with invertible 1x1 convolutions.
\newblock \emph{Advances in Neural Information Processing Systems 31}, pp.\
  10215--10224, 2018.

\bibitem[Kong \& Ping(2021)Kong and Ping]{kong2021fast}
Zhifeng Kong and Wei Ping.
\newblock On fast sampling of diffusion probabilistic models.
\newblock \emph{arXiv preprint arXiv:2106.00132}, 2021.

\bibitem[Kynk{\"a}{\"a}nniemi et~al.(2019)Kynk{\"a}{\"a}nniemi, Karras, Laine,
  Lehtinen, and Aila]{kynkaanniemi2019improved}
Tuomas Kynk{\"a}{\"a}nniemi, Tero Karras, Samuli Laine, Jaakko Lehtinen, and
  Timo Aila.
\newblock Improved precision and recall metric for assessing generative models.
\newblock \emph{Advances in Neural Information Processing Systems}, 32, 2019.

\bibitem[Li et~al.(2022{\natexlab{a}})Li, Yang, Chang, Chen, Feng, Xu, Li, and
  Chen]{li2022srdiff}
Haoying Li, Yifan Yang, Meng Chang, Shiqi Chen, Huajun Feng, Zhihai Xu, Qi~Li,
  and Yueting Chen.
\newblock {SRDiff}: Single image super-resolution with diffusion probabilistic
  models.
\newblock \emph{Neurocomputing}, 2022{\natexlab{a}}.

\bibitem[Li et~al.(2022{\natexlab{b}})Li, Thickstun, Gulrajani, Liang, and
  Hashimoto]{li2022diffusion}
Xiang Li, John Thickstun, Ishaan Gulrajani, Percy~S Liang, and Tatsunori~B
  Hashimoto.
\newblock Diffusion-{LM} improves controllable text generation.
\newblock \emph{Advances in Neural Information Processing Systems},
  35:\penalty0 4328--4343, 2022{\natexlab{b}}.

\bibitem[Liu et~al.(2021)Liu, Lin, Cao, Hu, Wei, Zhang, Lin, and
  Guo]{liu2021Swin}
Ze~Liu, Yutong Lin, Yue Cao, Han Hu, Yixuan Wei, Zheng Zhang, Stephen Lin, and
  Baining Guo.
\newblock Swin transformer: Hierarchical vision transformer using shifted
  windows.
\newblock \emph{arXiv preprint arXiv:2103.14030}, 2021.

\bibitem[Liu et~al.(2022)Liu, Mao, Wu, Feichtenhofer, Darrell, and
  Xie]{liu2022convnet}
Zhuang Liu, Hanzi Mao, Chao-Yuan Wu, Christoph Feichtenhofer, Trevor Darrell,
  and Saining Xie.
\newblock A convnet for the 2020s.
\newblock \emph{Proceedings of the IEEE/CVF Conference on Computer Vision and
  Pattern Recognition (CVPR)}, 2022.

\bibitem[Liu et~al.(2015)Liu, Luo, Wang, and Tang]{celebA2015deep}
Ziwei Liu, Ping Luo, Xiaogang Wang, and Xiaoou Tang.
\newblock Deep learning face attributes in the wild.
\newblock In \emph{Proceedings of the IEEE international conference on computer
  vision}, pp.\  3730--3738, 2015.

\bibitem[Loshchilov \& Hutter(2018)Loshchilov and Hutter]{2018adamw}
I.~Loshchilov and F.~Hutter.
\newblock Decoupled weight decay regularization.
\newblock In \emph{International Conference on Learning Representations}, 2018.

\bibitem[Lowe(2004)]{lowe2004distinctive}
David~G Lowe.
\newblock Distinctive image features from scale-invariant keypoints.
\newblock \emph{International journal of computer vision}, 60:\penalty0
  91--110, 2004.

\bibitem[Lu et~al.(2022)Lu, Zhou, Bao, Chen, Li, and Zhu]{lu2022dpm}
Cheng Lu, Yuhao Zhou, Fan Bao, Jianfei Chen, Chongxuan Li, and Jun Zhu.
\newblock {DPM-Solver}: A fast {ODE} solver for diffusion probabilistic model
  sampling in around 10 steps.
\newblock \emph{arXiv preprint arXiv:2206.00927}, 2022.

\bibitem[Luhman \& Luhman(2022)Luhman and Luhman]{luhman2022improving}
Troy Luhman and Eric Luhman.
\newblock Improving diffusion model efficiency through patching.
\newblock \emph{arXiv preprint arXiv:2207.04316}, 2022.

\bibitem[Luo(2022)]{luo2022understanding}
Calvin Luo.
\newblock Understanding diffusion models: A unified perspective.
\newblock \emph{arXiv preprint arXiv:2208.11970}, 2022.

\bibitem[Mairal et~al.(2007)Mairal, Elad, and Sapiro]{mairal2007sparse}
Julien Mairal, Michael Elad, and Guillermo Sapiro.
\newblock Sparse representation for color image restoration.
\newblock \emph{IEEE Transactions on image processing}, 17\penalty0
  (1):\penalty0 53--69, 2007.

\bibitem[Meng et~al.(2021)Meng, Song, Song, Wu, Zhu, and Ermon]{meng2021sdedit}
Chenlin Meng, Yang Song, Jiaming Song, Jiajun Wu, Jun-Yan Zhu, and Stefano
  Ermon.
\newblock Sdedit: Image synthesis and editing with stochastic differential
  equations.
\newblock \emph{arXiv preprint arXiv:2108.01073}, 2021.

\bibitem[Meng et~al.(2022)Meng, Gao, Kingma, Ermon, Ho, and
  Salimans]{meng2022distillation}
Chenlin Meng, Ruiqi Gao, Diederik~P Kingma, Stefano Ermon, Jonathan Ho, and Tim
  Salimans.
\newblock On distillation of guided diffusion models.
\newblock \emph{arXiv preprint arXiv:2210.03142}, 2022.

\bibitem[Mikolajczyk \& Schmid(2005)Mikolajczyk and
  Schmid]{mikolajczyk2005performance}
Krystian Mikolajczyk and Cordelia Schmid.
\newblock A performance evaluation of local descriptors.
\newblock \emph{IEEE transactions on pattern analysis and machine
  intelligence}, 27\penalty0 (10):\penalty0 1615--1630, 2005.

\bibitem[Nichol \& Dhariwal(2021)Nichol and Dhariwal]{nichol2021improved}
Alex Nichol and Prafulla Dhariwal.
\newblock Improved denoising diffusion probabilistic models.
\newblock \emph{arXiv preprint arXiv:2102.09672}, 2021.

\bibitem[Nichol et~al.(2021)Nichol, Dhariwal, Ramesh, Shyam, Mishkin, McGrew,
  Sutskever, and Chen]{nichol2021glide}
Alex Nichol, Prafulla Dhariwal, Aditya Ramesh, Pranav Shyam, Pamela Mishkin,
  Bob McGrew, Ilya Sutskever, and Mark Chen.
\newblock Glide: Towards photorealistic image generation and editing with
  text-guided diffusion models.
\newblock \emph{arXiv preprint arXiv:2112.10741}, 2021.

\bibitem[Pandey et~al.(2022)Pandey, Mukherjee, Rai, and
  Kumar]{pandey2022diffusevae}
Kushagra Pandey, Avideep Mukherjee, Piyush Rai, and Abhishek Kumar.
\newblock {DiffuseVAE}: Efficient, controllable and high-fidelity generation
  from low-dimensional latents.
\newblock \emph{arXiv preprint arXiv:2201.00308}, 2022.

\bibitem[Peebles \& Xie(2022)Peebles and Xie]{peebles2022scalable}
William Peebles and Saining Xie.
\newblock Scalable diffusion models with transformers.
\newblock \emph{arXiv preprint arXiv:2212.09748}, 2022.

\bibitem[Ramesh et~al.(2022)Ramesh, Dhariwal, Nichol, Chu, and
  Chen]{ramesh2022hierarchical}
Aditya Ramesh, Prafulla Dhariwal, Alex Nichol, Casey Chu, and Mark Chen.
\newblock Hierarchical text-conditional image generation with {CLIP} latents.
\newblock \emph{arXiv preprint arXiv:2204.06125}, 2022.

\bibitem[Rezende et~al.(2014)Rezende, Mohamed, and
  Wierstra]{rezende2014stochastic}
Danilo~Jimenez Rezende, Shakir Mohamed, and Daan Wierstra.
\newblock Stochastic backpropagation and approximate inference in deep
  generative models.
\newblock In \emph{Proceedings of the 31st International Conference on Machine
  Learning}, pp.\  1278--1286, 2014.

\bibitem[Robbins(1992)]{robbins1992empirical}
Herbert~E Robbins.
\newblock An empirical {B}ayes approach to statistics.
\newblock In \emph{Breakthroughs in Statistics: Foundations and basic theory},
  pp.\  388--394. Springer, 1992.

\bibitem[Rombach et~al.(2022)Rombach, Blattmann, Lorenz, Esser, and
  Ommer]{rombach2022high}
Robin Rombach, Andreas Blattmann, Dominik Lorenz, Patrick Esser, and Bj{\"o}rn
  Ommer.
\newblock High-resolution image synthesis with latent diffusion models.
\newblock In \emph{Proceedings of the IEEE/CVF Conference on Computer Vision
  and Pattern Recognition}, pp.\  10684--10695, 2022.

\bibitem[Ronneberger et~al.(2015)Ronneberger, P.Fischer, and
  Brox]{ronneberger2015unet}
O.~Ronneberger, P.Fischer, and T.~Brox.
\newblock {U-Net}: Convolutional networks for biomedical image segmentation.
\newblock In \emph{Medical Image Computing and Computer-Assisted Intervention
  (MICCAI)}, volume 9351 of \emph{LNCS}, pp.\  234--241. Springer, 2015.
\newblock URL
  \url{http://lmb.informatik.uni-freiburg.de/Publications/2015/RFB15a}.
\newblock (available on arXiv:1505.04597 [cs.CV]).

\bibitem[Saharia et~al.(2022)Saharia, Chan, Saxena, Li, Whang, Denton,
  Ghasemipour, Gontijo-Lopes, Ayan, Salimans, Ho, Fleet, and
  Norouzi]{saharia2022photorealistic}
Chitwan Saharia, William Chan, Saurabh Saxena, Lala Li, Jay Whang, Emily
  Denton, Seyed Kamyar~Seyed Ghasemipour, Raphael Gontijo-Lopes, Burcu~Karagol
  Ayan, Tim Salimans, Jonathan Ho, David~J. Fleet, and Mohammad Norouzi.
\newblock Photorealistic text-to-image diffusion models with deep language
  understanding.
\newblock In Alice~H. Oh, Alekh Agarwal, Danielle Belgrave, and Kyunghyun Cho
  (eds.), \emph{Advances in Neural Information Processing Systems}, 2022.
\newblock URL \url{https://openreview.net/forum?id=08Yk-n5l2Al}.

\bibitem[Salimans \& Ho(2022)Salimans and Ho]{salimans2022progressive}
Tim Salimans and Jonathan Ho.
\newblock Progressive distillation for fast sampling of diffusion models.
\newblock In \emph{International Conference on Learning Representations}, 2022.
\newblock URL \url{https://openreview.net/forum?id=TIdIXIpzhoI}.

\bibitem[San-Roman et~al.(2021)San-Roman, Nachmani, and Wolf]{san2021noise}
Robin San-Roman, Eliya Nachmani, and Lior Wolf.
\newblock Noise estimation for generative diffusion models.
\newblock \emph{arXiv preprint arXiv:2104.02600}, 2021.

\bibitem[Sohl-Dickstein et~al.(2015)Sohl-Dickstein, Weiss, Maheswaranathan, and
  Ganguli]{sohl2015deep}
Jascha Sohl-Dickstein, Eric Weiss, Niru Maheswaranathan, and Surya Ganguli.
\newblock Deep {U}nsupervised {L}earning {U}sing {N}onequilibrium
  {T}hermodynamics.
\newblock In \emph{International Conference on Machine Learning}, pp.\
  2256--2265, 2015.

\bibitem[Song et~al.(2020)Song, Meng, and Ermon]{song2020denoising}
Jiaming Song, Chenlin Meng, and Stefano Ermon.
\newblock Denoising diffusion implicit models.
\newblock \emph{arXiv preprint arXiv:2010.02502}, 2020.

\bibitem[Song \& Ermon(2019)Song and Ermon]{song2019generative}
Yang Song and Stefano Ermon.
\newblock Generative {M}odeling by {E}stimating {G}radients of the {D}ata
  {D}istribution.
\newblock In \emph{Advances in Neural Information Processing Systems}, pp.\
  11918--11930, 2019.

\bibitem[Song \& Ermon(2020)Song and Ermon]{song2020improved}
Yang Song and Stefano Ermon.
\newblock Improved {T}echniques for {T}raining {S}core-{B}ased {G}enerative
  {M}odels.
\newblock \emph{Advances in Neural Information Processing Systems}, 33, 2020.

\bibitem[Song et~al.(2021)Song, Sohl-Dickstein, Kingma, Kumar, Ermon, and
  Poole]{song2021scorebased}
Yang Song, Jascha Sohl-Dickstein, Diederik~P Kingma, Abhishek Kumar, Stefano
  Ermon, and Ben Poole.
\newblock Score-based generative modeling through stochastic differential
  equations.
\newblock In \emph{International Conference on Learning Representations}, 2021.
\newblock URL \url{https://openreview.net/forum?id=PxTIG12RRHS}.

\bibitem[Song et~al.(2023)Song, Dhariwal, Chen, and
  Sutskever]{song2023consistency}
Yang Song, Prafulla Dhariwal, Mark Chen, and Ilya Sutskever.
\newblock Consistency models.
\newblock \emph{arXiv preprint arXiv:2303.01469}, 2023.

\bibitem[Touvron et~al.(2022)Touvron, Cord, and Jegou]{Touvron2022DeiTIR}
Hugo Touvron, Matthieu Cord, and Herve Jegou.
\newblock {DeiT III}: Revenge of the {ViT}.
\newblock \emph{arXiv preprint arXiv:2204.07118}, 2022.

\bibitem[Tuytelaars \& Mikolajczyk(2008)Tuytelaars and
  Mikolajczyk]{tuytelaars2008local}
Tinne Tuytelaars and Krystian Mikolajczyk.
\newblock Local invariant feature detectors: a survey.
\newblock \emph{Foundations and trends{\textregistered} in computer graphics
  and vision}, 3\penalty0 (3):\penalty0 177--280, 2008.

\bibitem[Uria et~al.(2013)Uria, Murray, and Larochelle]{uria2013rnade}
Benigno Uria, Iain Murray, and Hugo Larochelle.
\newblock {RNADE}: The real-valued neural autoregressive density-estimator.
\newblock In \emph{Proceedings of the 26th International Conference on Neural
  Information Processing Systems-Volume 2}, pp.\  2175--2183, 2013.

\bibitem[Uria et~al.(2016)Uria, C{\^o}t{\'e}, Gregor, Murray, and
  Larochelle]{uria2016neural}
Benigno Uria, Marc-Alexandre C{\^o}t{\'e}, Karol Gregor, Iain Murray, and Hugo
  Larochelle.
\newblock Neural autoregressive distribution estimation.
\newblock \emph{The Journal of Machine Learning Research}, 17\penalty0
  (1):\penalty0 7184--7220, 2016.

\bibitem[Vahdat et~al.(2021)Vahdat, Kreis, and Kautz]{vahdat2021score}
Arash Vahdat, Karsten Kreis, and Jan Kautz.
\newblock Score-based generative modeling in latent space.
\newblock \emph{Advances in Neural Information Processing Systems},
  34:\penalty0 11287--11302, 2021.

\bibitem[Van Den~Oord et~al.(2016)Van Den~Oord, Kalchbrenner, and
  Kavukcuoglu]{oord2016pixel}
A\"{a}ron Van Den~Oord, Nal Kalchbrenner, and Koray Kavukcuoglu.
\newblock Pixel recurrent neural networks.
\newblock In \emph{Proceedings of the 33rd International Conference on
  International Conference on Machine Learning - Volume 48}, ICML'16, pp.\
  1747--1756. JMLR.org, 2016.
\newblock URL \url{http://dl.acm.org/citation.cfm?id=3045390.3045575}.

\bibitem[Vaswani et~al.(2017)Vaswani, Shazeer, Parmar, Uszkoreit, Jones, Gomez,
  Kaiser, and Polosukhin]{vaswani2017attention}
Ashish Vaswani, Noam Shazeer, Niki Parmar, Jakob Uszkoreit, Llion Jones,
  Aidan~N Gomez, {\L}ukasz Kaiser, and Illia Polosukhin.
\newblock Attention is all you need.
\newblock \emph{Advances in neural information processing systems}, 30, 2017.

\bibitem[Wang et~al.(2022)Wang, Zheng, He, Chen, and Zhou]{wang2022diffusion}
Zhendong Wang, Huangjie Zheng, Pengcheng He, Weizhu Chen, and Mingyuan Zhou.
\newblock {Diffusion-GAN}: Training {GANs} with diffusion.
\newblock \emph{International Conference on Learning Representations (ICLR)},
  2022.

\bibitem[Wang et~al.(2023)Wang, Jiang, Zheng, Wang, He, Wang, Chen, and
  Zhou]{wang2023patch}
Zhendong Wang, Yifan Jiang, Huangjie Zheng, Peihao Wang, Pengcheng He,
  Zhangyang Wang, Weizhu Chen, and Mingyuan Zhou.
\newblock Patch diffusion: Faster and more data-efficient training of diffusion
  models.
\newblock \emph{arXiv preprint arXiv:2304.12526}, 2023.

\bibitem[Xiao et~al.(2022)Xiao, Kreis, and Vahdat]{xiao2021tackling}
Zhisheng Xiao, Karsten Kreis, and Arash Vahdat.
\newblock Tackling the generative learning trilemma with denoising diffusion
  {GAN}s.
\newblock In \emph{International Conference on Learning Representations}, 2022.
\newblock URL \url{https://openreview.net/forum?id=JprM0p-q0Co}.

\bibitem[Yang et~al.(2022)Yang, Shih, Fu, Zhao, and Ji]{yang2022your}
Xiulong Yang, Sheng-Min Shih, Yinlin Fu, Xiaoting Zhao, and Shihao Ji.
\newblock Your {ViT} is secretly a hybrid discriminative-generative diffusion
  model.
\newblock \emph{arXiv preprint arXiv:2208.07791}, 2022.

\bibitem[Yu et~al.(2024)Yu, Simig, Flaherty, Aghajanyan, Zettlemoyer, and
  Lewis]{yu2024megabyte}
Lili Yu, D{\'a}niel Simig, Colin Flaherty, Armen Aghajanyan, Luke Zettlemoyer,
  and Mike Lewis.
\newblock Megabyte: Predicting million-byte sequences with multiscale
  transformers.
\newblock \emph{Advances in Neural Information Processing Systems}, 36, 2024.

\bibitem[Zhang \& Chen(2022)Zhang and Chen]{zhang2022fast}
Qinsheng Zhang and Yongxin Chen.
\newblock Fast sampling of diffusion models with exponential integrator.
\newblock \emph{arXiv preprint arXiv:2204.13902}, 2022.

\bibitem[Zhang et~al.(2018)Zhang, Isola, Efros, Shechtman, and
  Wang]{zhang2018perceptual}
Richard Zhang, Phillip Isola, Alexei~A Efros, Eli Shechtman, and Oliver Wang.
\newblock The unreasonable effectiveness of deep features as a perceptual
  metric.
\newblock In \emph{CVPR}, 2018.

\bibitem[Zheng et~al.(2022{\natexlab{a}})Zheng, Nie, Vahdat, Azizzadenesheli,
  and Anandkumar]{zheng2022fast}
Hongkai Zheng, Weili Nie, Arash Vahdat, Kamyar Azizzadenesheli, and Anima
  Anandkumar.
\newblock Fast sampling of diffusion models via operator learning.
\newblock \emph{arXiv preprint arXiv:2211.13449}, 2022{\natexlab{a}}.

\bibitem[Zheng et~al.(2023)Zheng, Nie, Vahdat, and Anandkumar]{zheng2023fast}
Hongkai Zheng, Weili Nie, Arash Vahdat, and Anima Anandkumar.
\newblock Fast training of diffusion models with masked transformers.
\newblock \emph{arXiv preprint arXiv:2306.09305}, 2023.

\bibitem[Zheng et~al.(2022{\natexlab{b}})Zheng, He, Chen, and
  Zhou]{zheng2022mixing}
Huangjie Zheng, Pengcheng He, Weizhu Chen, and Mingyuan Zhou.
\newblock Mixing and shifting: Exploiting global and local dependencies in
  vision mlps, 2022{\natexlab{b}}.

\bibitem[Zheng et~al.(2022{\natexlab{c}})Zheng, He, Chen, and
  Zhou]{zheng2022truncated}
Huangjie Zheng, Pengcheng He, Weizhu Chen, and Mingyuan Zhou.
\newblock Truncated diffusion probabilistic models and diffusion-based
  adversarial auto-encoders.
\newblock \emph{International Conference on Learning Representations (ICLR)},
  2022{\natexlab{c}}.

\bibitem[Zhou et~al.(2009)Zhou, Chen, Paisley, Ren, Sapiro, and
  Carin]{zhou2009non}
Mingyuan Zhou, Haojun Chen, John Paisley, Lu~Ren, Guillermo Sapiro, and
  Lawrence Carin.
\newblock Non-parametric {B}ayesian dictionary learning for sparse image
  representations.
\newblock \emph{Advances in neural information processing systems}, 22, 2009.

\bibitem[Zhou et~al.(2023)Zhou, Chen, Wang, and Zheng]{zhou2023beta}
Mingyuan Zhou, Tianqi Chen, Zhendong Wang, and Huangjie Zheng.
\newblock Beta diffusion.
\newblock In \emph{Neural Information Processing Systems}, 2023.
\newblock URL \url{https://arxiv.org/abs/2309.07867}.

\end{thebibliography}
\bibliographystyle{iclr2024_conference}

\clearpage
\appendix

\section{Discussions on related work}\label{sec:related}

\subsubsection*{More efficient diffusion models}

Diffusion models have garnered increasing attention from researchers seeking to enhance them in various ways. These efforts include combining variational loss with weighted denoising loss~\citep{dhariwal2021diffusion,kingma2021variational,karras2022edm}, accelerating sampling speed using ODE solvers~\citep{kong2021fast, san2021noise,song2020denoising,zhang2022fast,lu2022dpm}, and employing auxiliary models to improve training~\citep{rombach2022high,li2022diffusion,zheng2022truncated,pandey2022diffusevae}. Additionally, the iterative refinement property of diffusion models has inspired research in various domains, such as stable GAN training~\citep{xiao2021tackling,wang2022diffusion}, large-scale text-to-image models~\citep{kim2021diffusionclip,nichol2021glide,gu2021vector,saharia2022photorealistic}, and robust super-resolution and editing models~\citep{meng2021sdedit,choi2021ilvr,li2022srdiff}, among others.

With rapid advancements, diffusion models have often outperformed state-of-the-art generative adversarial networks (GANs, \citep{goodfellow2014generative}). GANs typically require adversarial training and are susceptible to issues such as unstable training and mode collapse. In contrast, diffusion models offer a versatile framework that is not bound by the traditional constraints associated with autoregressive models \citep{bengio1999modeling, uria2013rnade, uria2016neural, oord2016pixel}, variational autoencoders (VAEs, \citep{kingma2013auto, rezende2014stochastic}), or flow-based models \citep{dinh2014nice, dinh2016density, kingma2018glow, song2023consistency}.

While diffusion models have consistently demonstrated impressive performance, they face challenges related to substantial training and inference overheads, especially when applied to large high-resolution image datasets. Much of the research on diffusion models has focused on improving their sampling efficiency. Some studies~\citep{song2020denoising,lu2022dpm,karras2022edm} have refined sampling strategies and utilized sophisticated numerical solvers to enhance efficiency. In parallel, alternative approaches~\citep{salimans2022progressive,meng2022distillation,zheng2022fast,song2023consistency} have employed surrogate networks and distillation techniques. However, these solutions primarily target sampling efficiency and have not made substantial progress in reducing the training costs associated with diffusion models.

To address the challenge of training efficiency,  latent-space diffusion models were introduced to transform high-resolution images into a manageable, low-dimensional latent space \citep{vahdat2021score,rombach2022high}. \citet{ho2022cascaded} introduced a cascaded diffusion model approach, consisting of a foundational diffusion model for low-resolution images, followed by super-resolution ones. Similarly, \citet{wang2023patch} proposed a segmented training paradigm, conducting score matching at a patch level. 
While this U-Net-based approach exhibits certain resemblances to ours, wherein a set of smaller patches serves as input for the Transformer blocks, notable distinctions exist.

Recently, \citet{bao2022all,bao2023one} and \citet{peebles2022scalable} explored the use of Transformer architecture for diffusion models instead of the conventional U-Net architecture. Additionally, masked Transformers were utilized 
to improve the training convergence of diffusion models \citep{gao2023masked,zheng2022fast}. However, that approach, which involves processing both masked and unmasked tokens, substantially increases the training overhead per iteration. In contrast, our method prioritizes efficiency by exclusively utilizing the unmasked tokens during training.

\subsubsection*{Backbone of diffusion models}

A crucial element within diffusion models revolves around the network architecture employed in their iterative refinement-based generative process. The initial breakthrough, which propelled score- or diffusion-based generative models into the spotlight, can be attributed to the integration of the convolutional operation-based U-Net architecture \citep{song2019generative,ho2020denoising}. Originally conceived by \citet{ronneberger2015unet} for biomedical image segmentation, the U-Net design incorporates both a progressive downsampling path and a progressive upsampling path, enabling it to generate images of the same dimensions as its input. In the context of diffusion models, the U-Net often undergoes further modifications, such as the incorporation of attention blocks, residual blocks, and adaptive normalization layers  \citep{dhariwal2021diffusion}.
 
While the downsampling and upsampling convolutional layers in U-Net allow it to effectively capture local structures, achieving a good understanding of global structures often requires a deep architecture and self-attention modules embodied in the architecture, resulting in high computational complexity during both training and generation. 
Instead of depending on a large number of convolutional layers, a classical approach in signal and image processing involves the extraction of local feature descriptors and their application in downstream tasks \citep{lowe2004distinctive,mikolajczyk2005performance,tuytelaars2008local}. For instance, techniques like dictionary learning and sparse coding operate on the principle that each local patch can be expressed as a sparse linear combination of dictionary atoms learned from image patches. This learned patch dictionary can then be used to enrich all overlapping patches for various image-processing tasks, such as denoising and inpainting \citep{aharon2006k,mairal2007sparse,zhou2009non}.

The emergence of ViT in image classification \citep{dosovitskiy2021an}, which integrates extracting feature embeddings from non-overlapping local patches with the global attention mechanism of the Transformer, marks a resurgence of this classical local-feature-based concept. ViT has not only competed effectively with state-of-the-art convolution-based networks in both image classification \citep{Touvron2022DeiTIR} and generation \citep{esser2021taming}, but has also been adapted to replace the U-Net in diffusion models \citep{yang2022your,bao2022all,peebles2022scalable,gao2023masked}, achieving state-of-the-art image generation performance. This underscores the notion that the convolutional architecture can be supplanted by the enrichment of local features and their orchestration through global attention. 
ViT-based models start by enriching patch-level representations and then utilize a sequence of Transformer blocks to capture dependencies among local patches. They often demand smaller patch sizes to capture fine-grained local details and a large number of attention layers to capture global dependencies, which results in a demand of enormous data to fit. This necessitates lengthy sequence computations and deep networks to achieve optimal results, thereby incurring significant computational costs. 
To enhance efficiency and expedite convergence, recent works have explored hybridizing the entangled global-local structure and multi-scale properties from convolutional networks into hierarchical transformers
\citep{liu2021Swin,zheng2022mixing,liu2022convnet,yu2024megabyte}. 

To bridge the gap between the convolutional U-Nets and the Transformer architecture, LEGO tackles the diffusion modeling from a view of local-feature enrichment and global-content orchestration. Inspired by convolutional networks, LEGO enriches local features at both local and global scales. However, different from a series of downsampling and upsampling stages used in the U-Nets, LEGO leverages varied patch decomposition methods. This distinction 
results in diverse feature representations. To achieve a global-content orchestration, LEGO deploys self-attention, whose length varies depending on the `LEGO brick'. In some cases, attention is confined to specific image patches defined by the brick size, while in some other bricks, it extends over the entire image. This selective mechanism allows for more focused and efficient feature aggregation.

\section{Limitations and Future Work}
The current work exhibits several limitations. Firstly, we have not explored the application of LEGO bricks for text-guided image generation. Secondly, in generating class-conditional images at resolutions much higher than the training data, categories with low diversity may produce results that lack photorealism. Thirdly, our primary focus has been on stacking LEGO bricks either in a progressive-growth or progressive-refinement manner, with other stacking strategies remaining relatively unexplored. Fourthly, our selection of which LEGO bricks to skip at different time steps during generation relies on heuristic criteria, and there is room for developing more principled skipping strategies.

While the use of LEGO bricks has led to considerable computational savings in both training and generation, it still requires state-of-the-art GPU resources that are very expensive to run and hence typically beyond the reach of budget-sensitive projects. For example, training LEGO diffusion with 512M images on ImageNet $512\times 512$ took 32 NVIDIA A100 GPUs with about 14 days. 

It's important to note that like other diffusion-based generative models, LEGO diffusion can potentially be used for harmful purposes when trained on inappropriate image datasets. Addressing this concern is a broader issue in diffusion models, and LEGO bricks do not appear to inherently include mechanisms to mitigate such risks.

\section{\rev{Experimental settings}}\label{sec:experiment-detail}

\textbf{Datasets: }
We conduct experiments using the CelebA dataset~\citep{celebA2015deep}, which comprises 162,770 images of human faces, and the ImageNet dataset~\citep{deng2009imagenet}, consisting of 1.3 million natural images categorized into 1,000 classes. For data pre-processing, we follow the convention to first apply a center cropping and then resize it to $64\times 64$~\citep{song2021scorebased} on CelebA; on ImageNet, we pre-process images at three different resolutions: $64\times 64$, $256\times 256$, and $512\times 512$.

For images of resolution $64\times 64$, we directly train the model in the pixel space. For higher-resolution images, we follow \citet{rombach2022high} to first encode the images into a latent space with a down-sampling factor of 8 and then train a diffusion model on that latent space. Specifically, 
we employ the EMA checkpoint of autoencoder from Stable Diffusion\footnote{\url{https://huggingface.co/stabilityai/sd-vae-ft-mse-original}} to pre-process the images of size $256\times 256\times 3$ into $32\times 32 \times 4$, and those of size $512\times 512\times 3$ into $64\times 64 \times 4$. 

\textbf{Training settings: } %
For our LEGO model backbones, we employ EDM~\citep{karras2022edm} preconditioning on pixel-space and iDDPM~\citep{nichol2021improved,dhariwal2021diffusion}  on latent-space data, strictly adhering to the diffusion designs prescribed by these respective preconditioning strategies. 

During training, we employ the AdamW optimizer~\citep{2018adamw} for all experiments. For DDPM preconditioning, we use a fixed learning rate of $1\times 10^{-4}$; for EDM preconditioning, we adopt the proposed learning rate warmup strategy, where the learning rate increases linearly to $1\times 10^{-4}$ until iterated with 10k images. The batch size per GPU is set at 64 for all experiments. When trained on CelebA, we finish the training when the model is iterated with 2M images, and when trained on ImageNet, including $64\times64$, $256\times256$, and $512\times512$ resolution, we finish the training when the model is iterated with 512M images.

\textbf{Evaluation: } %
For performance evaluation, we primarily use FLOPs to measure the generation %
cost and generate 50,000 random images to compute the FID score as a performance metric. Additionally, %
we include the metrics~\citep{ding2022continuous,kynkaanniemi2019improved} employed by~\citet{peebles2022scalable} for a comprehensive comparison. In the context of image generation, we employ the Heun sampler when using EDM preconditioning~\citep{karras2022edm} and the DDPM sampler when utilizing iDDPM preconditioning~\citep{nichol2021improved}.

\rev{
Specifically, we employed the same suite, namely the ADM Tensorflow evaluation suite implemented in \citet{dhariwal2021diffusion}\footnote{{\small 
\url{https://github.com/openai/guided-diffusion/tree/main/evaluations}}}. We also utilized their pre-extracted features for the reference batch to ensure that our evaluations are rigorous and reliable. Additionally, we tested the evaluation with scripts and reference batch provided in EDM \citep{karras2022edm} \footnote{\rev{\url{https://github.com/nvlabs/edm}}}. We observed that the results from both evaluation suites were consistent, showing no significant differences (up to two decimal places).
}

\paragraph{Diffusion settings} We deploy our LEGO model backbones with two commonly-used parameterization, namely DDPM~\citep{nichol2021improved,dhariwal2021diffusion} and EDM~\citep{karras2022edm}. Recall the training loss of diffusing models in \Eqref{eq:dpm training loss}:
\begin{align}
&\textstyle \rvtheta^\star = \argmin_\rvtheta \mathop{\mathbb{E}}_{t, \rvx_t, \rvepsilon} [\lambda_t \| \rvepsilon_{\rvtheta}(\rvx_t,t) - \rvepsilon \|_2^2],~\text{or}~ \rvtheta^\star = \argmin_\rvtheta \mathop{\mathbb{E}}_{t, \rvx_t, \rvepsilon} [\lambda_t^\prime \| \hat{\rvx}_0(\rvx_t,t;{\rvtheta}) - \rvx_0 \|_2^2]. \notag
\end{align}
The DDPM preconditioning follows the former equation, training the model to predict the noise injected to $\rvx_t$. Following~\citet{peebles2022scalable}, we use the vanilla DDPM~\citep{ho2020denoising} loss, where $\lambda_t = 1$, and use the linear diffusion schedule. The EDM preconditioning can be regarded as an SDE model optimized using the latter equation, where the output is parameterized with:
$$
\hat{\rvx}_0(\rvx_t,t;{\rvtheta}) = f_\theta(\rvx_t, t;{\rvtheta}) = c_{\text {skip }}(\sigma) \rvx_t + c_{\text {out }}(\sigma) \epsilon_\theta\left(c_{\text {in }}(\sigma) \rvx_t ; c_{\text {noise }}(\sigma)\right),
$$
where $\sigma$ depends on the noise-level of the current timestep. $c_{\text {skip }}(\sigma)$, $c_{\text {in }}$, $c_{\text {out }}$, and $c_{\text {noise }}$ are 4 hyper-parameters depending on $\sigma$. We strictly follow EDM to set these parameters and use the EDM schedule. More details regarding the diffusion settings can be found in \citet{ho2020denoising} and \citet{karras2022edm}.

During generation, for DDPM preconditioning, we use the diffusion sampler for 250 steps with a uniform stride, proposed in \citet{nichol2021improved}. For EDM preconditioning, we use the 2$^{nd}$ order Heun sampler in \citet{karras2022edm}. On CelebA, we sample 75 steps with the deterministic sampler. On Imagenet, we sample for 256 steps with the stochastic sampler. The stochasticity-related hyper-parameters are set as: $S_\text{churn}=10$, $S_\text{min}=0.05$, $S_\text{max}=20$, and $S_\text{noise}=1.003$.

\begin{table*}[t]
\setlength{\tabcolsep}{2.1pt}
\footnotesize
    \centering
    \caption{LEGO-Diffusion model architectures with different configurations. Following the convention, we introduce three different configurations---Small, Large, and XLarge---for different model capacities. The generation resolution is assumed to be $64\times64$. {In each LEGO brick, $l$ and $d$ represent the size of local receptive fields and their embedding dimension, respectively. The number of tokens (attention span) in a LEGO brick  is determined as the brick size (input \& output patch size) $r\times r$ divided by its local-receptive field size $\ell\times \ell$.}  LEGO-PG stacks the three LEGO bricks with brick sizes of $4\times4$, $16\times16$, and $64\times64$ from the bottom to the top to form the network backbone for diffusion modeling. By contrast, LEGO-PR stacks them in the reverse order. }%
    \label{tab:model_config}
    \resizebox{\linewidth}{!}{
    \begin{tabular}{c|c|c|c|c}
    \toprule[1.5pt]
              {Brick Size} & Layer Name & LEGO-S & LEGO-L & LEGO-XL \\
            \midrule
    \multirow{5}{*}{$4\times4$}  & Token Embedding  &  ${\ell}=2;d=384$ & ${\ell}=2;d=1024$ & ${\ell}=4;d=1152$ \\
    \cmidrule{2-5}
                                 &  \makecell{LEGO\\patch-brick} & \multicolumn{1}{c|}{$\left[\begin{array}{c}
                                   \text{\# attention heads} = 6 \\
                                   \text{mlp ratio} = 4
                              \end{array}\right] \times 2$} &
\multicolumn{1}{c|}{$\left[\begin{array}{c}
                                   \text{\# attention heads} = 16 \\
                                   \text{mlp ratio} = 4
                              \end{array}\right] \times 4$} &
\multicolumn{1}{c}{$\left[\begin{array}{c}
                                   \text{\# attention heads} = 16 \\
                                   \text{mlp ratio} = 4
                              \end{array}\right] \times 4$} 
                              \\
        \midrule
    \multirow{5}{*}{$ 16\times 16$} & Token Embedding & ${\ell}=8;d=384$ & ${\ell}=8;d=1024$ & ${\ell}=8;d=1152$ \\
    \cmidrule{2-5}
              & \makecell{LEGO\\patch-brick} & \multicolumn{1}{c|}{$\left[\begin{array}{c}
                                   \text{\# attention heads} = 6 \\
                                   \text{mlp ratio} = 4
                              \end{array}\right] \times 4$} &
\multicolumn{1}{c|}{$\left[\begin{array}{c}
                                   \text{\# attention heads} = 16 \\
                                   \text{mlp ratio} = 4
                              \end{array}\right] \times 8$} &
\multicolumn{1}{c}{$\left[\begin{array}{c}
                                   \text{\# attention heads} = 16 \\
                                   \text{mlp ratio} = 4
                              \end{array}\right] \times 12$}                             \\
        \midrule
    \multirow{5}{*}{$ 64\times 64$} & Token Embedding & ${\ell}=2;d=384$ & ${\ell}=2;d=1024$ & ${\ell}=2;d=1152$\\
        \cmidrule{2-5}
             & \makecell{LEGO\\image-brick} & \multicolumn{1}{c|}{$\left[\begin{array}{c}
                                   \text{\# attention heads} = 6 \\
                                   \text{mlp ratio} = 4
                              \end{array}\right] \times 6$} &
\multicolumn{1}{c|}{$\left[\begin{array}{c}
                                   \text{\# attention heads} = 16 \\
                                   \text{mlp ratio} = 4
                              \end{array}\right] \times 12$} &
\multicolumn{1}{c}{$\left[\begin{array}{c}
                                   \text{\# attention heads} = 16 \\
                                   \text{mlp ratio} = 4
                              \end{array}\right] \times 14$} \\ \cmidrule{1-5}
\multicolumn{2}{c|}{ \rev{Parameter size (M)}} &  \rev{35} & \rev{464} &  \rev{681} \\
            \bottomrule[1.5pt]
    \end{tabular}}
\end{table*}

\begin{table*}[t]
\setlength{\tabcolsep}{2.1pt}
\footnotesize
    \centering
    \caption{{Analogous to Table~\ref{tab:model_config} for LEGO to model $32\times 32 $ resolution.}
    }
    \label{tab:model_config_32}
    \resizebox{\linewidth}{!}{
    \begin{tabular}{c|c|c|c|c}
    \toprule[1.5pt]
              {Brick Size} & Layer Name & LEGO-S & LEGO-L & LEGO-XL \\
            \midrule
    \multirow{5}{*}{$4\times4$}  & Token Embedding  &  ${\ell}=2;d=384$ & ${\ell}=2;d=1024$ & ${\ell}=4;d=1152$ \\
    \cmidrule{2-5}
                                 &  \makecell{LEGO\\patch-brick} & \multicolumn{1}{c|}{$\left[\begin{array}{c}
                                   \text{\# attention heads} = 6 \\
                                   \text{mlp ratio} = 4
                              \end{array}\right] \times 2$} &
\multicolumn{1}{c|}{$\left[\begin{array}{c}
                                   \text{\# attention heads} = 16 \\
                                   \text{mlp ratio} = 4
                              \end{array}\right] \times 4$} &
\multicolumn{1}{c}{$\left[\begin{array}{c}
                                   \text{\# attention heads} = 16 \\
                                   \text{mlp ratio} = 4
                              \end{array}\right] \times 4$} 
                              \\
        \midrule
    \multirow{5}{*}{$ 8\times 8$} & Token Embedding & ${\ell}=4;d=384$ & ${\ell}=4;d=1024$ & ${\ell}=4;d=1152$ \\
    \cmidrule{2-5}
              & \makecell{LEGO\\patch-brick} & \multicolumn{1}{c|}{$\left[\begin{array}{c}
                                   \text{\# attention heads} = 6 \\
                                   \text{mlp ratio} = 4
                              \end{array}\right] \times 4$} &
\multicolumn{1}{c|}{$\left[\begin{array}{c}
                                   \text{\# attention heads} = 16 \\
                                   \text{mlp ratio} = 4
                              \end{array}\right] \times 8$} &
\multicolumn{1}{c}{$\left[\begin{array}{c}
                                   \text{\# attention heads} = 16 \\
                                   \text{mlp ratio} = 4
                              \end{array}\right] \times 12$}                             \\
        \midrule
    \multirow{5}{*}{$ 32\times 32$} & Token Embedding & ${\ell}_3=2;d=384$ & ${\ell}_3=2;d=1024$ & ${\ell}_3=2;d=1152$\\
        \cmidrule{2-5}
             & \makecell{LEGO\\image-brick} & \multicolumn{1}{c|}{$\left[\begin{array}{c}
                                   \text{\# attention heads} = 6 \\
                                   \text{mlp ratio} = 4
                              \end{array}\right] \times 6$} &
\multicolumn{1}{c|}{$\left[\begin{array}{c}
                                   \text{\# attention heads} = 16 \\
                                   \text{mlp ratio} = 4
                              \end{array}\right] \times 12$} &
\multicolumn{1}{c}{$\left[\begin{array}{c}
                                   \text{\# attention heads} = 16 \\
                                   \text{mlp ratio} = 4
                              \end{array}\right] \times 14$} \\ \cmidrule{1-5}
\multicolumn{2}{c|}{ \rev{Parameter size (M)}} &  \rev{35} & \rev{464} &  \rev{681} \\
            \bottomrule[1.5pt]
    \end{tabular}}
\end{table*}

\paragraph{Model design}
{
We provide an overview of the architecture configurations for the LEGO models with different capacities in Tables \ref{tab:model_config} and \ref{tab:model_config_32}. Since Table~\ref{tab:model_config_32} only differs from Table~\ref{tab:model_config} in the brick sizes and local receptive field sizes, we will focus on explaining Table~\ref{tab:model_config} in detail. 
Table~\ref{tab:model_config} consists of three LEGO bricks, with two of them being patch-level bricks and one being an image-level brick. When using the PG-based spatial refinement approach, we stack the bricks from the bottom to the top following the order: $4\times 4 \rightarrow 16\times 16 \rightarrow 64\times 64$. In other words, the $4\times 4$ patch-level brick is placed at the bottom, followed by the $16\times 16$ patch-level brick, and then the $64\times 64$ image-level brick. Conversely, for the PR-based spatial refinement approach, we stack the bricks from the bottom to the top following the order: $64\times 64 \rightarrow 16\times 16 \rightarrow 4\times 4$.
}
Adhering to conventional modeling techniques, we've outlined three distinct configurations --- Small (LEGO-S), Large (LEGO-L), and XLarge (LEGO-XL) --- used in our main experiments to cater to varying model capacities. 
{Across bricks, the main differences are in token embedding and the number of attention heads, which revolves around the feature size of a local receptive field. Each brick specifies its patch size  $r_k\times r_k$, local receptive field size $\ell_k\times \ell_k$, and embedding dimension (number of channels) $d_k$. The number of tokens (attention span) is determined as $(r_k/\ell_k)^2$.
}
{
In each brick, the configurations for LEGO-S, LEGO-L, and LEGO-XL also differ in terms of the number of attention heads and the number of DiT blocks used in the current brick to ensure scalability and adaptability across the architecture. To maintain generation coherence, we set the local-receptive-field size of the image-level brick as $2\times 2$, which effectively turns it into a DiT with a long attention span, albeit with significantly fewer layers compared to standard DiT models. Moreover, the image-level brick can be safely skipped at appropriate time steps during generation.
}
We find the proposed configuration strikes a favorable balance between computational efficiency and performance, which is comprehensively studied in our ablation studies on the effects of LEGO bricks in Appendix~\ref{sec:ablation}. \rev{During training, the LEGO patch-bricks can operate on sampled patches to further save memory and computation. In our experiments on the CelebA and ImageNet datasets, we utilize 50\% and 75\% of all non-overlapping patches for training patch-bricks, respectively.} As noted in Section \ref{sec:method-tech},  there are instances where $\hat{\rvx}_{0,(i,j)}^{(k-1)}$ contains missing values as some patches are not sampled in the preceding brick. In such cases, we replace these missing values with the corresponding pixels from $\rvx_0$ to ensure that the required $\hat{\rvx}_{0,(i,j)}^{(k-1)}$ is complete.

During training, there are a couple of viable approaches. One method is to stack all bricks and train them end-to-end.
Alternatively, a brick-by-brick %
training methodology can be deployed. In our practical studies, we did not find clear performance differences within these choices.

\section{Additional Experiment Results}
\label{app:results}
\subsection{Skipping LEGO bricks in spatial refinement: comparing PG and PR strategies }

Here we provide the qualitative results to support the analysis of the brick skipping mechanism of LEGO shown in Section \ref{sec:skip_sampling}.
Progressive growth (PG) and progressive refinement (PR) offer two distinct spatial refinement strategies in the LEGO model, and leveraging their unique properties to skip corresponding LEGO bricks during sampling presents a non-trivial challenge. We visualize the differences between these strategies in Figures \ref{fig:stage-vis-pg} and \ref{fig:stage-vis-pr}. In PG, the model initially uses patch-bricks to generate patches and then employs an image-brick to {re-aggregate these local features.}
Conversely, PR begins by utilizing the image-brick to establish a global structure and then employs local feature-oriented patch-bricks to refine details on top of it. These differences become more evident when $t$ is larger.
Based on these observations, we propose that LEGO-PG can skip its top-level image-brick responsible for handling global image information when $t$ is small, while LEGO-PR skips its top-level patch-brick, which is responsible for processing $4\times4$ local patch information, when $t$ is large.

\begin{figure}[ht]
    \centering
    \includegraphics[width=.95\textwidth]{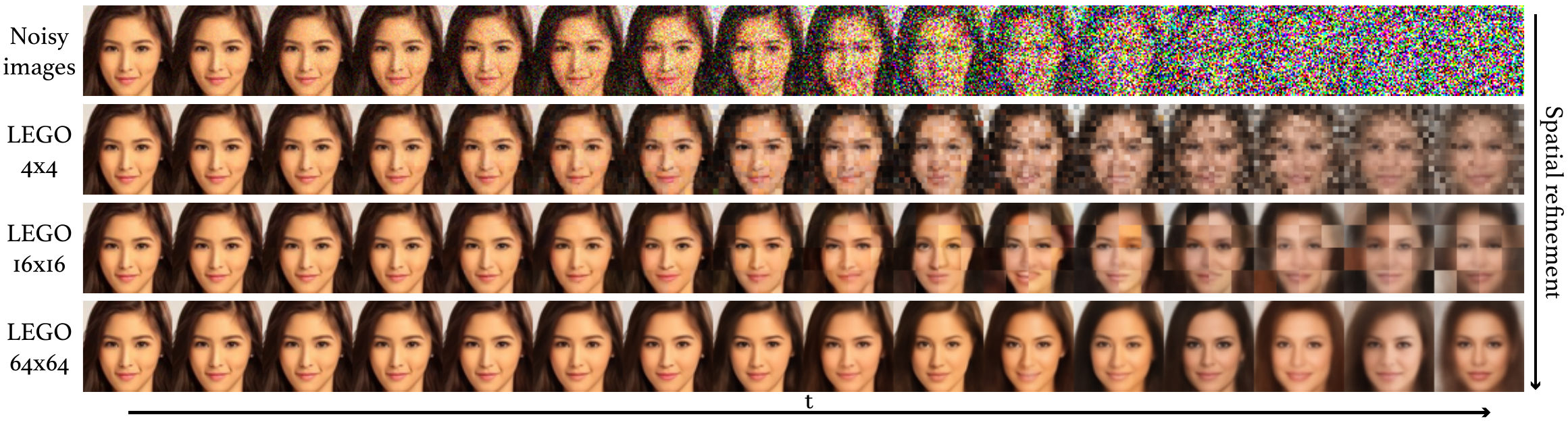}
    \vspace{-4mm}
    \caption{Visualization of the spatial 
    refinement process using the LEGO-PG model: each column represents $\hat{\rvx}_0^{(k)}$ at different timesteps obtained by corrupting the clean image, shown in the top row of the first column, at various noise levels. Within each column, the spatial refinement process, as described in \eqref{eq:lego_module}, proceeds from the top row to the bottom row. %
    }
    \label{fig:stage-vis-pg}
\end{figure}

\begin{figure}[ht]
    \centering
    \includegraphics[width=.97\textwidth]{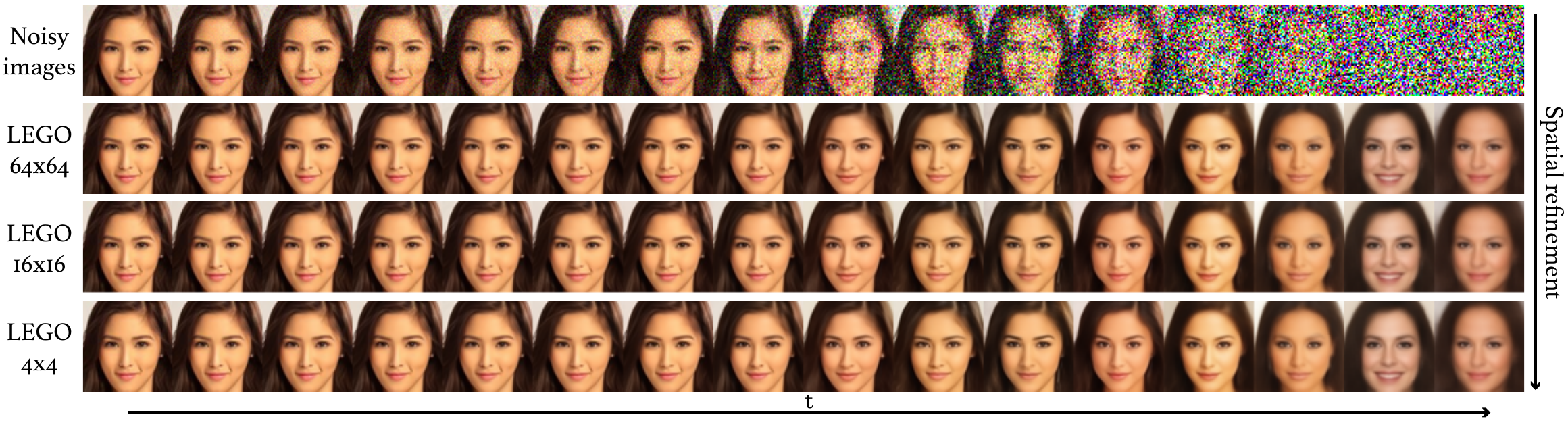}
    \vspace{-4mm}
    \caption{Analogous results of \Figref{fig:stage-vis-pg}, using the LEGO-PR model.}
    \label{fig:stage-vis-pr}
\end{figure}

Below we show analogous results of \Figref{fig:sampling_skip_IN256} on different datasets, illustrated in Figures \ref{fig:sampling_skip_celeba}-\ref{fig:sampling_skip_IN512}.

\begin{figure}[h]
\centering
\includegraphics[width=\textwidth]{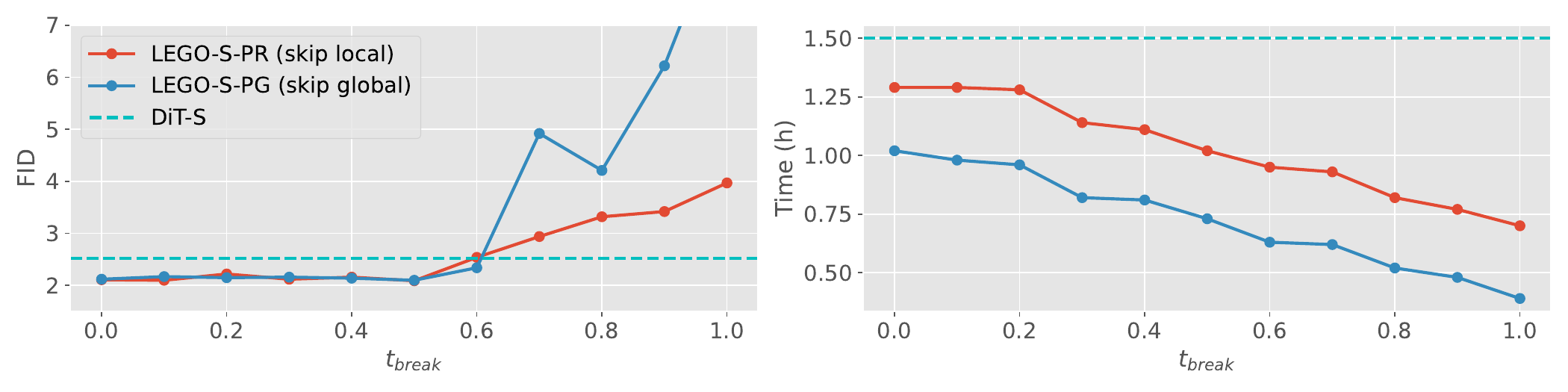}\vspace{-5mm}
\caption{Analogous results to \Figref{fig:sampling_skip_IN256}. The experiments are conducted using the LEGO-S model with eight NVIDIA A100 GPUs on the CelebA ($64\times 64$) dataset.}
\label{fig:sampling_skip_celeba}
\end{figure}
\begin{figure}[h]
\centering
\includegraphics[width=\textwidth]{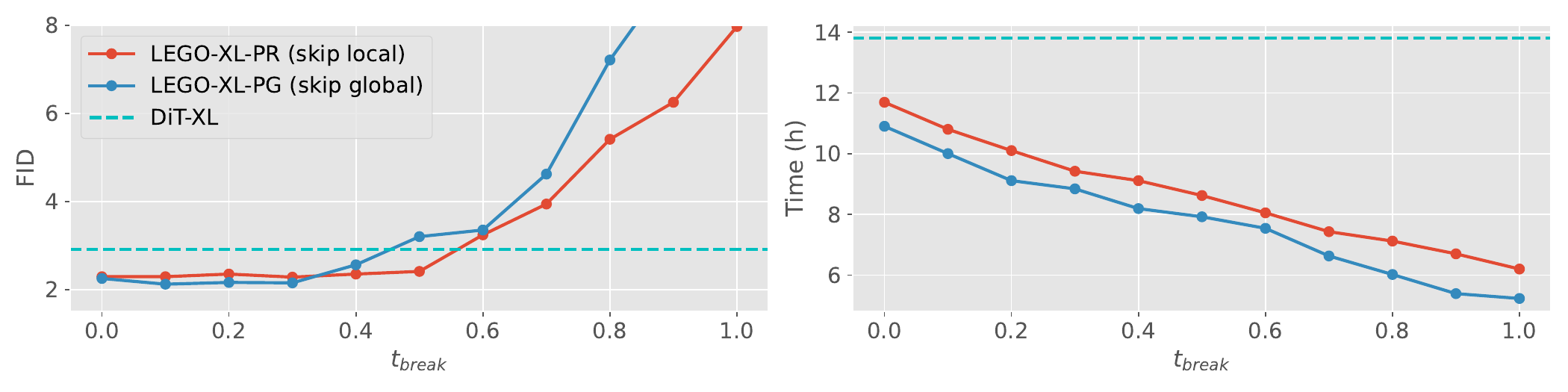}\vspace{-5mm}
\caption{\rev{Analogous results to \Figref{fig:sampling_skip_IN256}, conducted using the LEGO-L model on the ImageNet ($64\times 64$) dataset with classifier-free guidance. }}
\label{fig:sampling_skip_IN64}
\end{figure}
\begin{figure}[h]
\centering
\includegraphics[width=\textwidth]{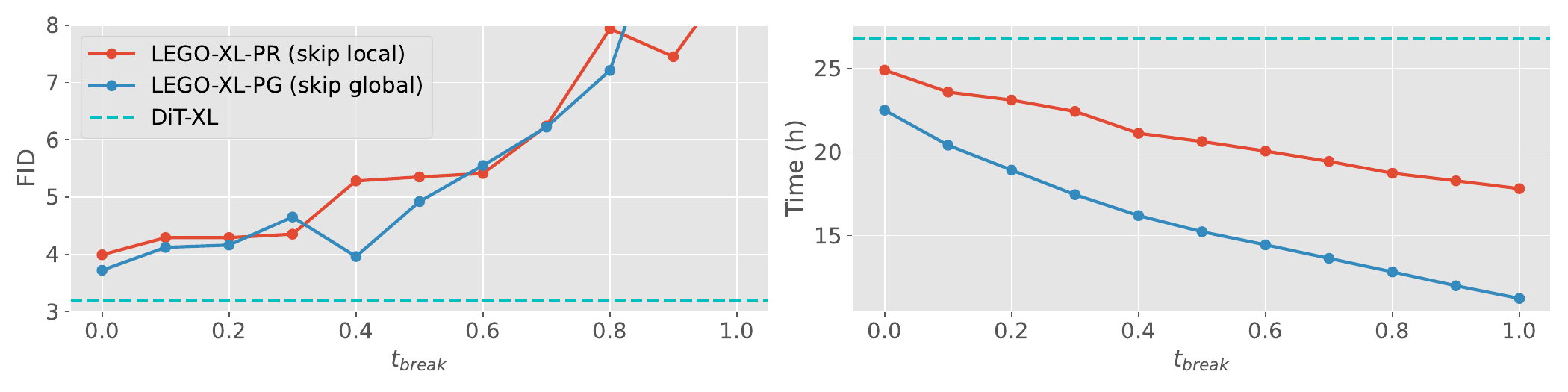}
\caption{\rev{Analogous results to \Figref{fig:sampling_skip_IN256}, conducted using the LEGO-XL model on the ImageNet ($512\times 512$) dataset with classifier-free guidance. }}
\label{fig:sampling_skip_IN512}
\end{figure}

\subsection{Ablation Studies: On the effects of LEGO bricks}\label{sec:ablation}
In this section, we first study the effects of different LEGO bricks. Note that the {patch-bricks} target to provide fine-grained details in generation and cost fewer computation resource as they take smaller input resolution {and shorter attention span}; the image-brick ensures the global structure of the generation and consistency within local regions, but requires more computation cost. Therefore, to find %
{a good} 
trade-off between generative performance and computation cost, we conduct experiments to find {the right} balance between the {patch-bricks and image-brick.}

\paragraph{Proportion of layers assigned to the image-brick} 

\begin{wrapfigure}{r}{0.55\textwidth}
\begin{center}\vspace{-6mm}
   \includegraphics[width=0.55\textwidth]{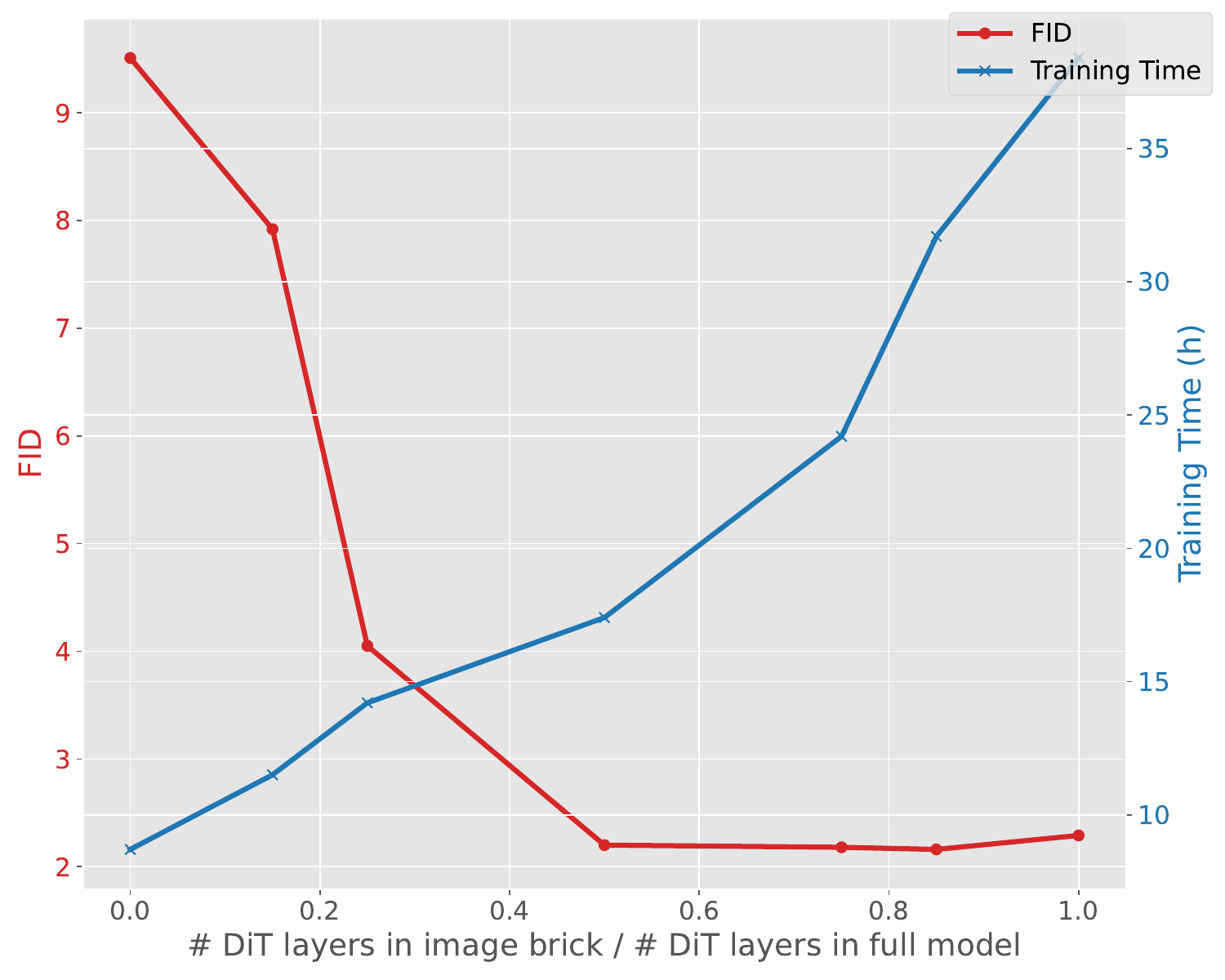}
\end{center}\vspace{-6mm}
        \caption{{Visualization: how the FID and training time change as the ratio of the depth (number of DiT-based Transformer layers) of the image-brick to that of the patch-brick increases from 0 to 1.}}
        \label{fig:local_global_ablation}
\end{wrapfigure}
We first study the trade-off between the training time and the ratio of the depth (number of DiT block layers) of the image-brick to the overall depth of the LEGO model. For simplicity, we consider a LEGO-Diffusion model consisting of a patch-brick with a patch size of $4\times 4$ and an image-brick with a patch size of $64\times 64$, both with a receptive-field-size set as $2\times 2$. We also let both bricks have the same token embedding dimension, number of attention heads, MLP ratio, \textit{etc}. Fixing the total depth as 12, we vary the depth assigned to the image-brick from 0 to 12 and show the results in  \Figref{fig:local_global_ablation}. For a LEGO-Diffusion model consisting of a patch-brick with a patch size of $4\times 4$ and an image-brick with a patch size of $64\times 64$, both with a receptive field size set as $2\times 2$, we visualize how the FID and training time change as the ratio of the depth (number of DiT-based Transformer layers) of the image-brick to that of the patch-brick increases from 0 to 1. Specifically, at the beginning of x-axis, we have a patch-brick with 12 DiT-based Transformer layers and an image-brick with 0 DiT-based Transformer layers, and at the end, we have a patch-brick with 0 DiT-based Transformer layers and an image-brick with 12 DiT-based Transformer layers. We can observe that when we allocate the majority of layers to the patch-brick, the training is efficient but results in a relatively higher FID. As we assign more layers to the image-brick, the FID improves, but the model requires longer training time. Empirically, we find that when the ratio is greater than 50\%, the FID improvement becomes less significant, while efficiency decreases at the same time. This observation motivates us to allocate approximately 50\% of the layers to the patch-brick.

\begin{figure}[h]
        \includegraphics[width=\textwidth]{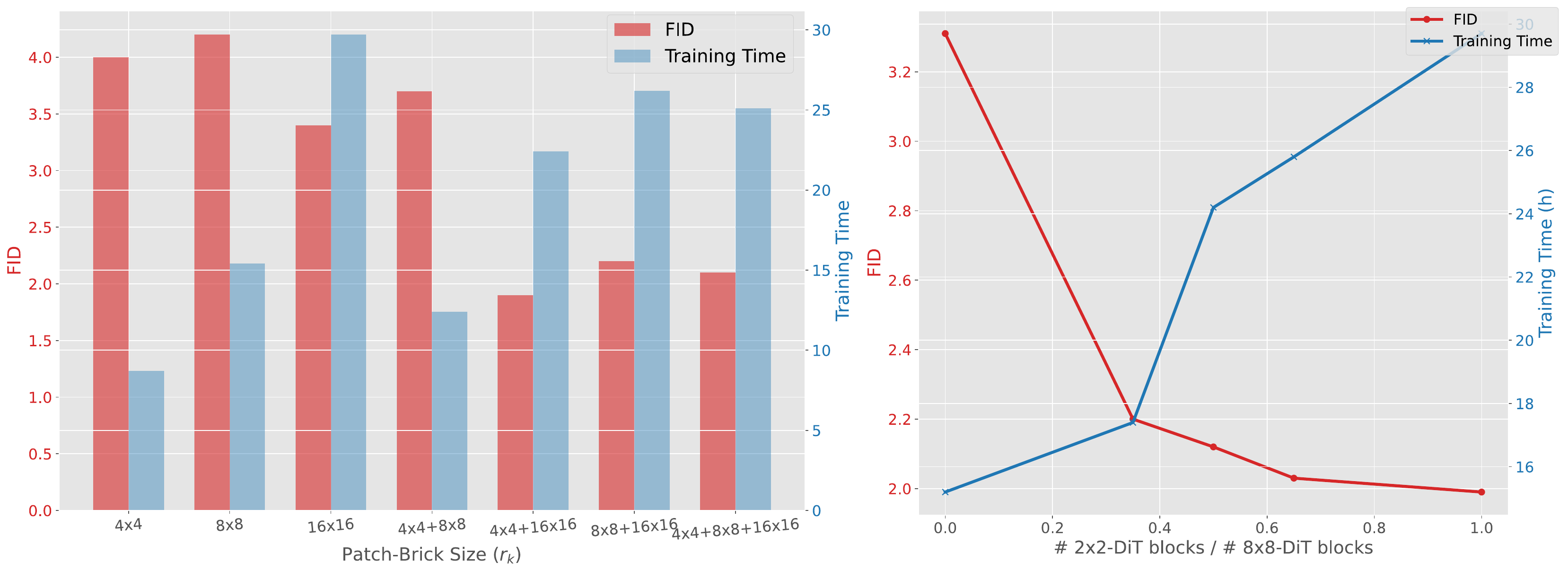}\vspace{-4mm}
        \caption{
        {Ablation study on the construction of the patch-bricks when the image-brick is fixed of six layers. (\textit{Left:}) Comparison of three different model configurations: using a single patch-brick with six layers at three different brick sizes, using two patch-bricks, each with three layers, at three different combinations of brick sizes, and using three patch-bricks, each with two layers, each at different patch sizes. (\textit{Right:}) 
        Consider a model comprising two patch-bricks: one with a brick size of $4\times 4$ and a local receptive field of $2\times 2$, and the other with a brick size of $16\times 16$ and a local receptive field of $8\times 8$. We visualize how the FID and training time change as we vary the ratio of the number of layers assigned to the $4\times 4$ patch brick to the total number of layers of these two bricks combined, while keeping the total number of layers in the two bricks fixed at six.
        }}
        \label{fig:local_ablation}
\end{figure}

\paragraph{Multi-scale local bricks} 
{Inspired by the multi-stage design commonly found in convolutional networks, we investigate the impact of using multi-scale patch-bricks on our diffusion model's performance. To isolate the effect, we first fix the configurations of both patch- and image-bricks, including the local-receptive-field size that is fixed at $2\times 2$ and the total number of DiT-block layers is fixed at 12, varying only the resolution of the patches used in training the patch-bricks. Specifically, we experiment with brick sizes of $4\times4$, $8\times8$, $16\times16$, as well as combinations of them. The corresponding FID scores and training times are summarized on the left panel of Figure~\ref{fig:local_ablation}. Our findings indicate that when using a single patch-brick, a $16 \times 16$ brick size yields the best performance but incurs higher training time due to the need for longer token embedding sequences. Using a combination of $4 \times 4$ and $8 \times 8$ patch-bricks does not significantly improve the FID score, possibly because $8 \times 8$ patches do not add substantially more information than $4 \times 4$ patches. However, when a $16 \times 16$ patch-brick is combined with a smaller patch-brick, we see a significant improvement in FID. Notably, the combination of $4 \times 4$ and $16 \times 16$ patch-bricks offers the best trade-off between FID and training time. As a result, this combination of patch sizes is employed when modeling $64\times 64$ resolutions in our experiments.
}

{
To further optimize training efficiency, for the $16 \times 16$ patch-brick, we increase its receptive field size from $2\times 2$ to $8\times 8$ and investigate how to assign the six DiT block layers between the $4\times 4$ patch-brick and the $16\times 16$ patch-brick. The results shown on the right panel of Figure~\ref{fig:local_ablation}  reveal that assigning more layers to the $4\times 4$ patch-brick in general results in lower FID but longer training time. 
Finally, we observe that when the depth of the $16 \times 16$ patch-brick is approximately twice that of the $4 \times 4$ patch-brick, the model achieves a satisfactory balance between performance and efficiency. %
These observations have been leveraged in the construction of the LEGO model configurations, which are outlined in Tables~\ref{tab:model_config} and \ref{tab:model_config_32}. It's worth noting that additional tuning of configurations, such as employing specific local receptive field sizes, adjusting the embedding dimension of the local receptive field, varying the number of attention heads, and so on for each LEGO brick, holds the potential for further performance improvements. However, we leave these aspects for future exploration and research.
}

\subsection{Additional Generation results}
\textbf{Generating beyond training resolution: } Below we first provide additional panorama image generations at resolutions significantly higher than the training images, as shown in Figures \ref{fig:paranoma}-\ref{fig:paranoma_512}. For all panorama generations (including the top panel in \Figref{fig:imagenet-vis}), we utilize class-conditional LEGO-PG, trained on ImageNet ($256 \times 256$ and $512 \times 512$). \rev{ As deployed in \citet{bar2023multidiffusion}, large content is achievable through a pre-trained text-to-image diffusion model trained on images of lower resolution. In our study, we adopt this methodology to show the efficacy of the LEGO framework in such tasks. Notably, the LEGO framework exhibits enhanced capabilities in managing spatial dependencies, which is a critical aspect in this context. For each paranoma generation, we first initialize the latent representation of the panorama image with white Gaussian noise at the desired resolution, such as $(1280/8)\times (256/8)\times 4$ for generating $1280\times 256$ RGB images in Figure \ref{fig:paranoma}. During the generation process, at each timestep, we employ a sliding stride of 7 and have LEGO predict noise within a window of $(256/8) \times (256/8)$ until the target resolution is covered.  The predicted noise at each latent voxel is averaged based on the number of times it is generated, taking into account overlaps. When generating images or patterns with a mixture of class conditions, the idea is to produce diverse outputs that adhere to different class characteristics within a single canvas. This could be achieved by dividing the canvas into different regions, where each region corresponds to a specific class condition. In our evaluation, we employed the LPIPS distance metric proposed by~\citet{zhang2018perceptual}, to assess the similarity between sections of panoramic generations and images from our training set. For each image, we use either a grid-based cropping or a random cropping process. Our analysis involved 100 generated images from various categories. The results are presented in Table~\ref{tab:lpips_comparison}, including both the average and the standard deviation of the LPIPS distances. Notably, when employing the random cropping method, the quality of panoramas generated using the LEGO model evidently exceeded that of the DiT model. This difference is particularly located in areas comprising outputs from multiple diffusion models, highlighting LEGO's capacity in managing spatial relationships among image~patches.}

\begin{table}[th]
  \centering
  \caption{\rev{Comparision of large-content generation using LPIPS distance (lower is better). }} \label{tab:lpips_comparison}
\setlength{\tabcolsep}{1.0mm}{ 
\scalebox{1.0}{\color{black}
\begin{tabular}{l|c|c|c|c}
\toprule[1.5pt]
Crop    & Grid - 256   & Random - 256 & Grid - 512   & Random - 512 \\\midrule
DiT     & 0.15 $\pm$ 0.07 & 0.66 $\pm$ 0.13 & 0.55 $\pm$ 0.07 & 0.76 $\pm$ 0.06 \\ \midrule
LEGO-PG & 0.14 $\pm$ 0.07 & 0.21 $\pm$ 0.17 & 0.36 $\pm$ 0.07 & 0.37 $\pm$ 0.06 \\
LEGO-PR & 0.15 $\pm$ 0.07 & 0.25 $\pm$ 0.15 & 0.55 $\pm$ 0.07 & 0.55 $\pm$ 0.05 \\
\bottomrule[1.5pt]
\end{tabular}
}}
\end{table}

\begin{figure}
    \centering
    \includegraphics[width=\textwidth]{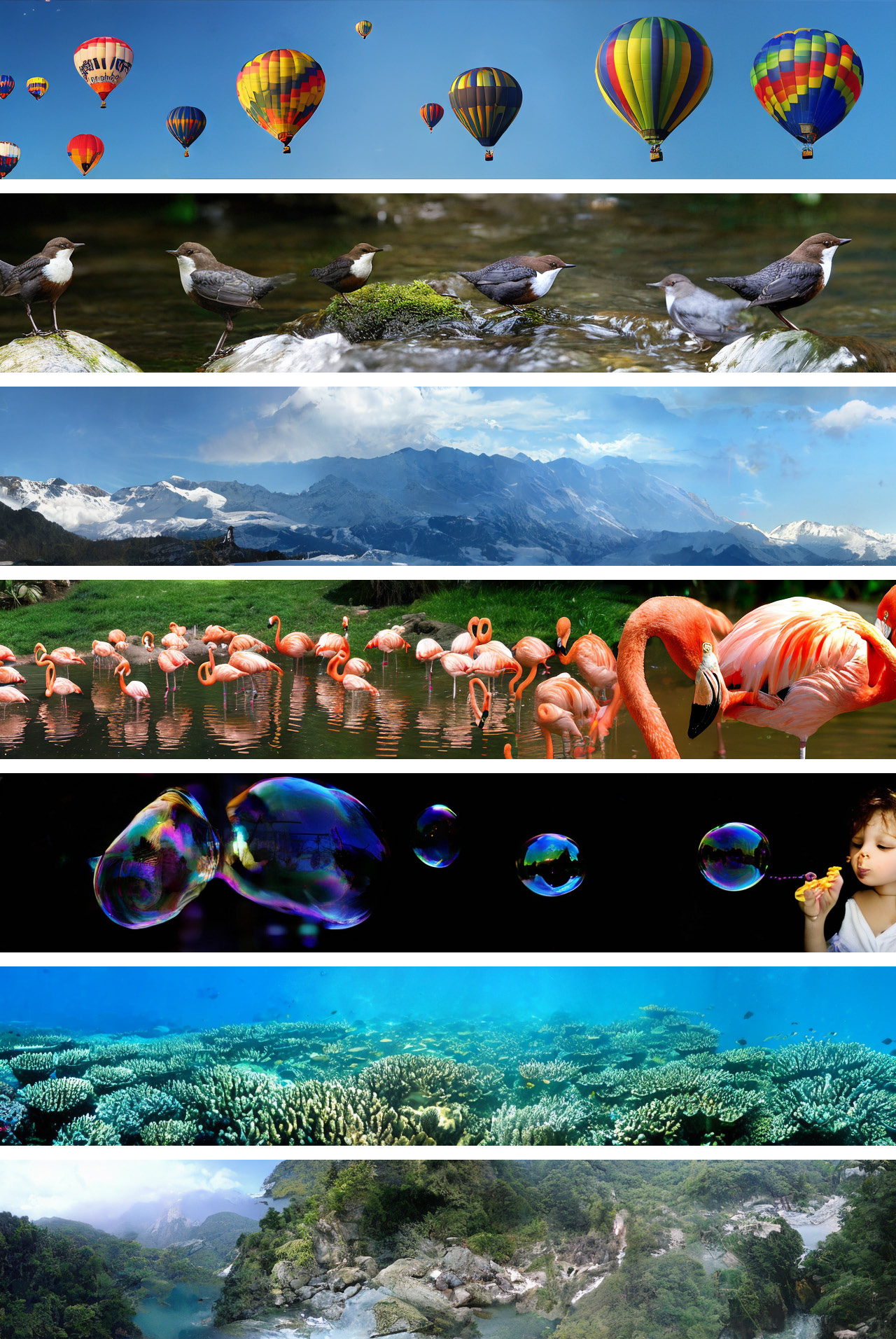}
    \caption{Paranoma generation ($1280 \times 256$) with class-conditional LEGO-PG, trained on ImageNet ($256\times 256$). (\textit{From top to bottom}) Class: Balloon (index 417), Water ouzel (index 20), Alp (index 970), Flamingo (index 130), Bubble (index 971), Coral reef (index 973), and Valley (index 979). }
    \label{fig:paranoma}
\end{figure}

\begin{figure}[h]
    \centering
    \includegraphics[width=\textwidth]{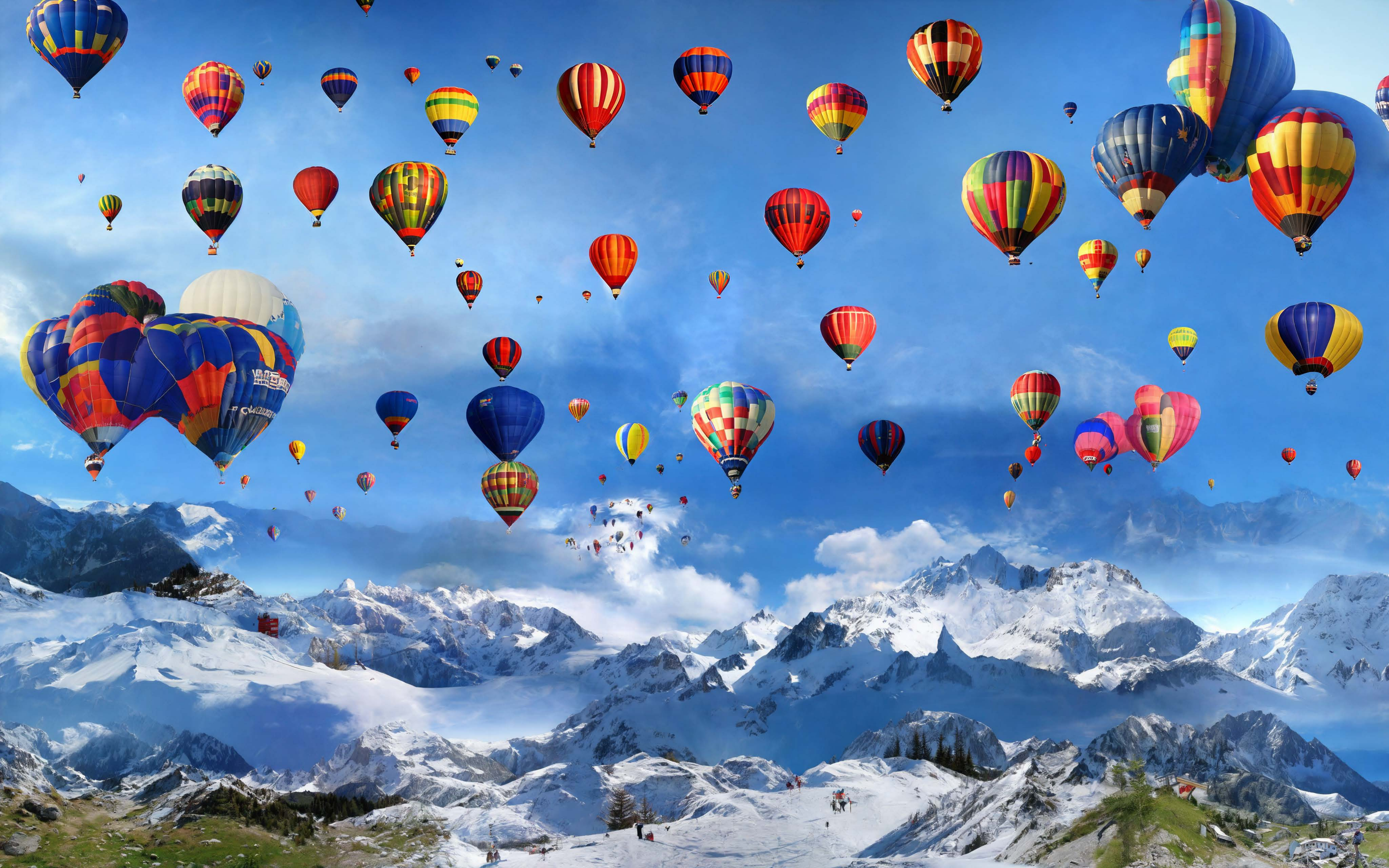}
    \caption{\rev{Paranoma generation of 4K ($3840  \times 2160$) resolution with a mixture of class conditions, produced with class-conditional LEGO-PG, trained on ImageNet ($512\times 512$). }}    
    \label{fig:4k}
\end{figure}

\begin{figure}
    \centering
    \includegraphics[width=\textwidth]{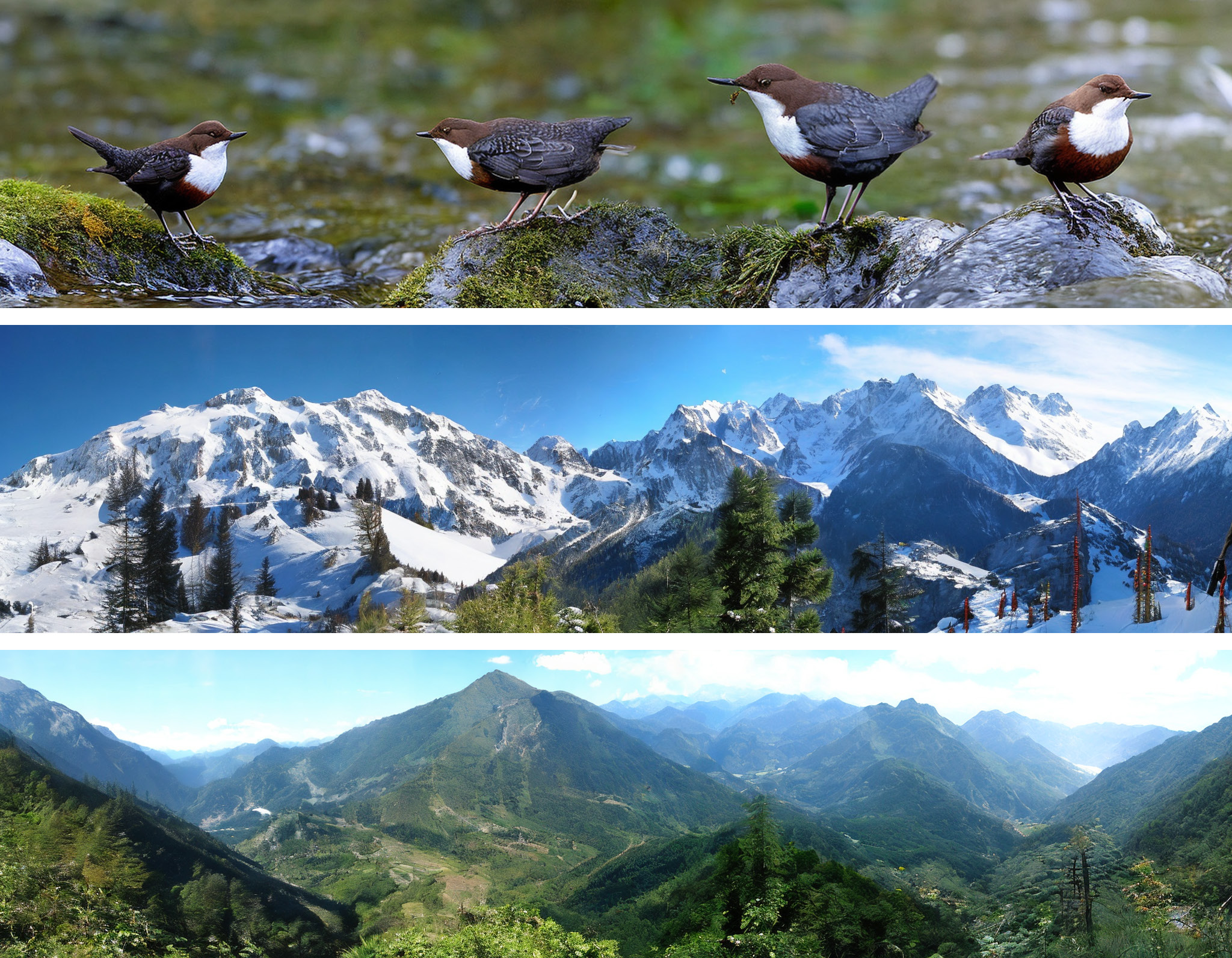}
    \caption{\rev{Paranoma generation ($2048 \times 512$) with class-conditional LEGO-PG, trained on ImageNet ($512\times 512$). (\textit{From top to bottom}) Class: Water ouzel (index 20), Alp (index 970), and Valley (index 979). }}
    \label{fig:paranoma_512}
\end{figure}

\begin{figure}
    \centering
    \includegraphics[width=\textwidth]{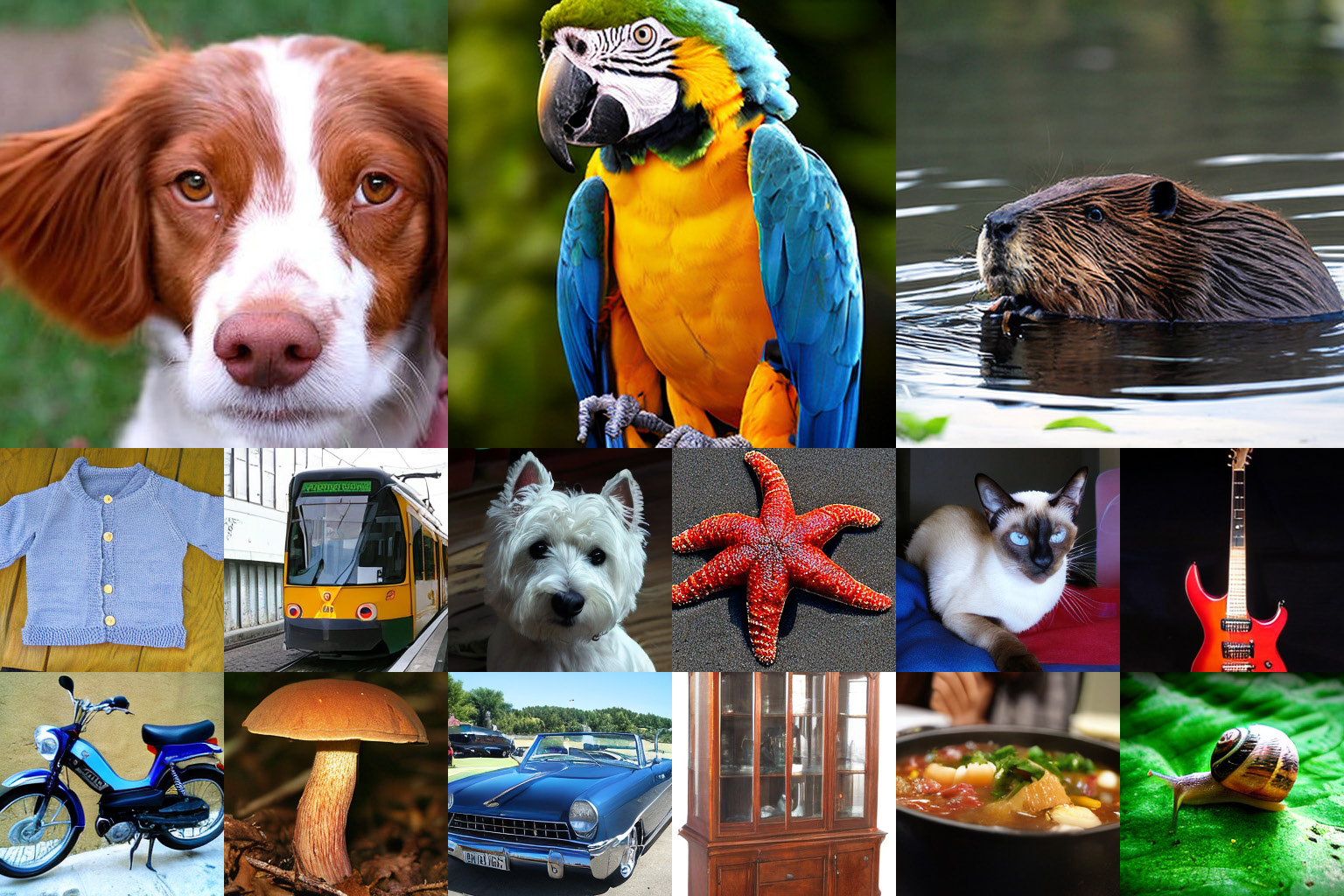}
    \caption{Randomly generated images, using LEGO-PG (cfg-scale=4.0)}
    \label{fig:addition_gen_1}
\end{figure}

\begin{figure}
    \centering
    \includegraphics[width=\textwidth]{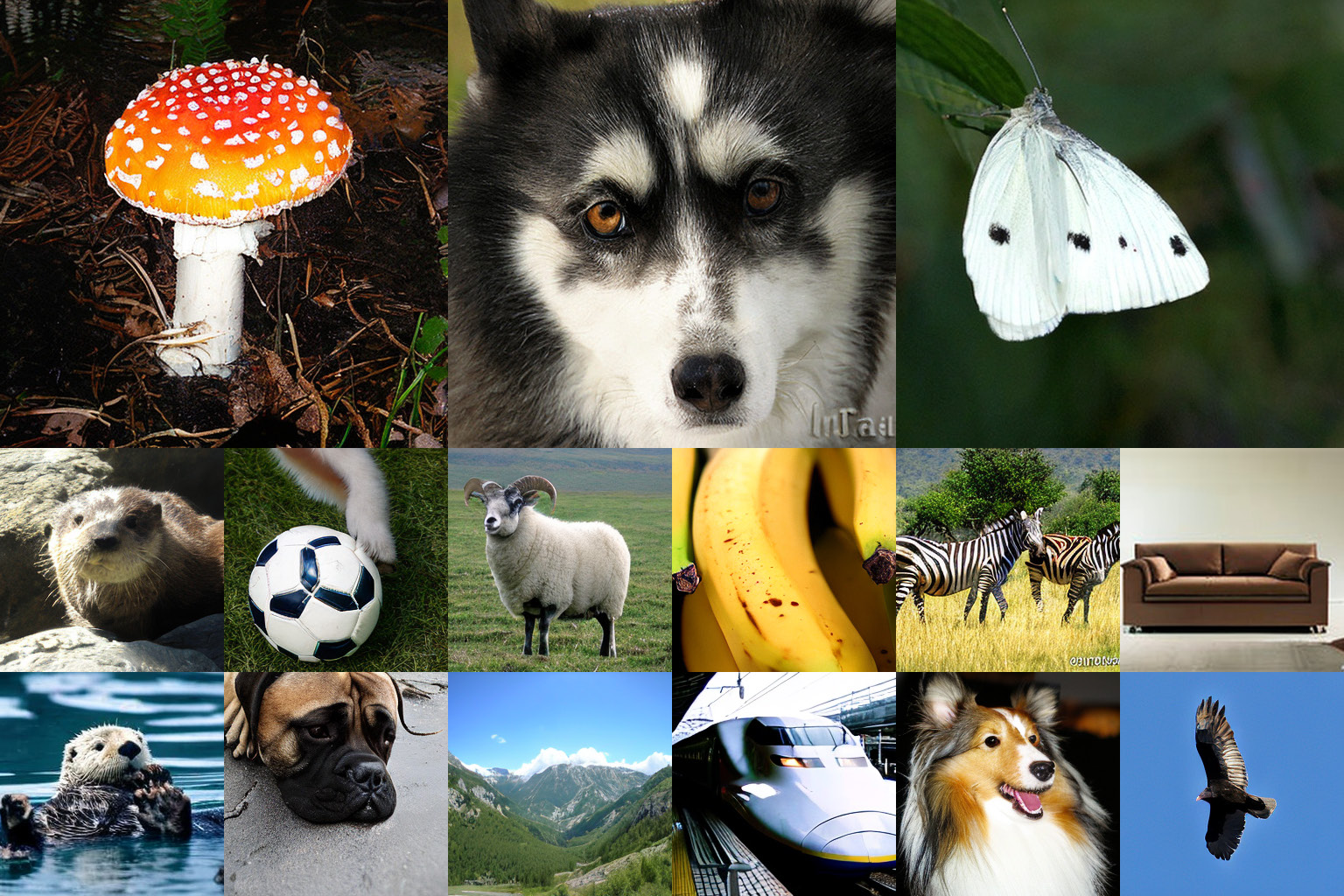}
    \caption{Randomly generated images, using LEGO-PG (cfg-scale=1.5)}
    \label{fig:addition_gen_2}
\end{figure}

\begin{figure}
    \centering
    \includegraphics[width=\textwidth]{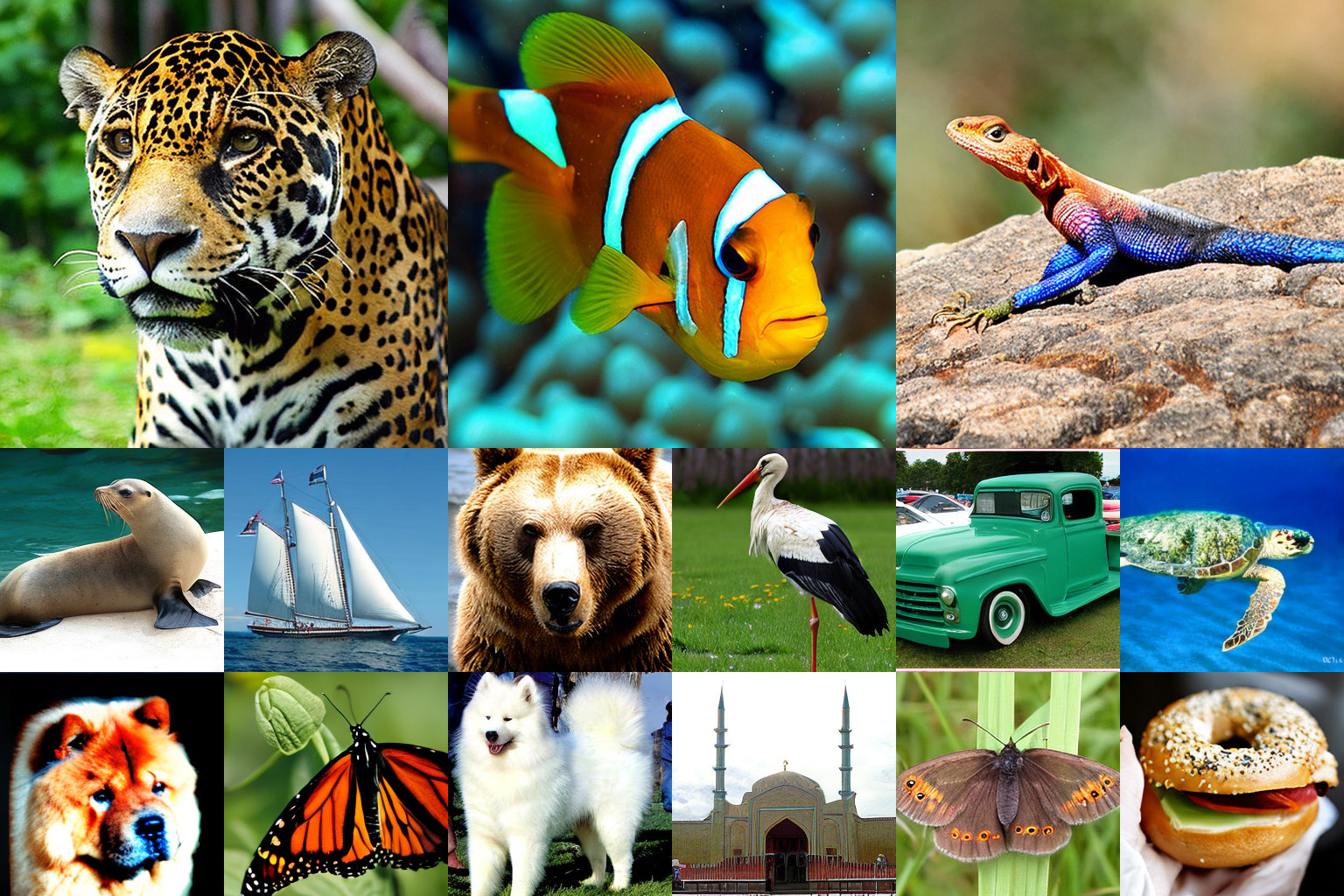}
    \caption{Randomly generated images, using LEGO-PR (cfg-scale=4.0)}
    \label{fig:addition_gen_1_pr}
\end{figure}

\begin{figure}
    \centering
    \includegraphics[width=\textwidth]{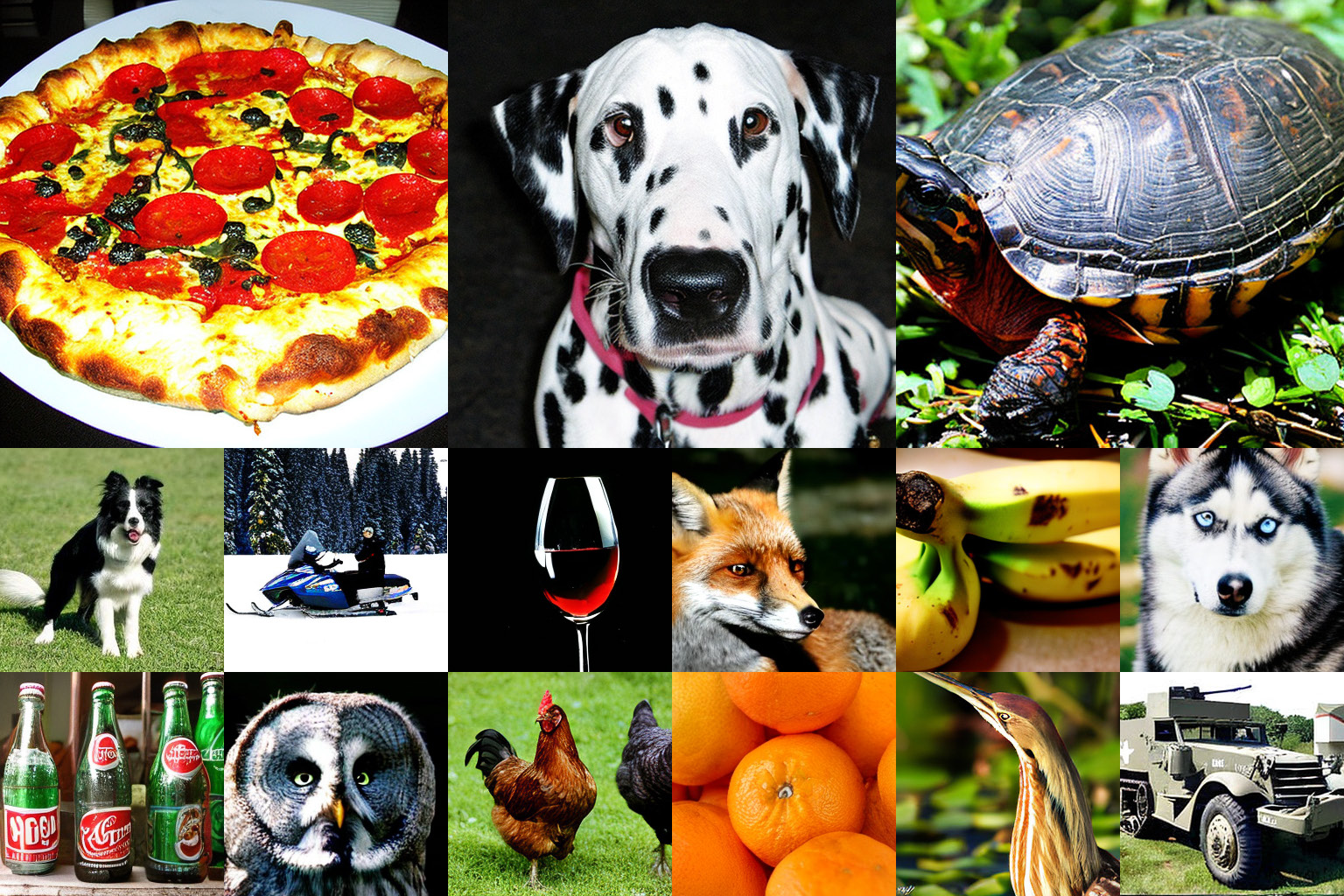}
    \caption{Randomly generated images, using LEGO-PR (cfg-scale=1.5)}
    \label{fig:addition_gen_2_pr}
\end{figure}
\end{document}